\documentclass[review,authoryear]{elsarticle}
\usepackage[natbibapa]{apacite}
\usepackage{bm}
\usepackage{subfig}
\usepackage{float}
\usepackage{multirow}
\usepackage{scalerel}
\usepackage{amsmath,amssymb,amsthm}
\usepackage{lineno,hyperref}
\newtheorem{theorem}{Theorem}
\newtheorem{lemma}[theorem]{Lemma}
\modulolinenumbers[5]
\journal{Journal of \LaTeX\ Templates}
\begin{document}

\begin{frontmatter}

\title{Tree-structured multi-stage principal component analysis (TMPCA): theory and applications}

\author[USC]{Yuanhang Su\corref{cor1}}\ead{suyuanhang@googlemail.com}
\author[USC]{Ruiyuan Lin}\ead{ruiyuanl@usc.edu}
\author[USC]{C.-C. Jay Kuo}\ead{cckuo@sipi.usc.edu}
\cortext[cor1]{Corresponding Author\\ 
Phone: +1-213-281-2388}
\address[USC]{University of Southern California, Ming Hsieh
Department of Electrical Engineering\\ 3740 McClintock Avenue, Los
Angeles, CA, United States}

\begin{abstract}

A PCA based sequence-to-vector (seq2vec) dimension reduction method for the text classification problem, called the tree-structured multi-stage principal component analysis (TMPCA) is presented in this paper. Theoretical analysis and applicability of TMPCA are demonstrated as an extension to our previous work \citep{TMPCA_ICPR}. Unlike
conventional word-to-vector embedding methods, the TMPCA method conducts
dimension reduction at the sequence level without labeled training data.
Furthermore, it can preserve the sequential structure of input
sequences. We show that TMPCA is computationally efficient and able to
facilitate sequence-based text classification tasks by preserving strong
mutual information between its input and output mathematically.  It is
also demonstrated by experimental results that a dense (fully connected)
network trained on the TMPCA preprocessed data achieves better
performance than state-of-the-art fastText and other
neural-network-based solutions. 

\end{abstract}


\end{frontmatter}

%\linenumbers

\section{Introduction}\label{sec:intro}

In natural language processing (NLP), dimension reduction is often
required to alleviate the so-called ``curse of dimensionality" problem. 
This occurs when the numericalized input data are in a
sparse high-dimensional space \citep{Sparsity}. Such a problem partly arises from the
large size of vocabulary and partly comes from the sentence variations
with similar meanings. Both contribute to high-degree data pattern
diversity, and a high dimensional space is required to represent the
data in a numerical form adequately.  Due to the ever-increasing data in
the Internet nowadays, the language data become even more diverse.  As a
result, previously well-solved problems such as text classification (TC)
face new challenges \citep{Conv_text, overview_TC}. An effective
dimension reduction technique remains to play a critical role in
tackling these challenges. The new dimension reduction solution should satisfy 
the following criteria:

\begin{itemize}
\item Reduce the input dimension
\item Retain the input information
\end{itemize}

More specifically, dimension reduction technique should maximally preserve 
the input information given the limited dimension available for representing 
the input data. Different classifiers will perform differently given the 
same input data. Our objective is not to find such best performing classifiers, 
but to propose a dimension reduction technique that can facilitate the following 
classification process.

There are many ways to reduce the language data to a compact form.  The
most popular ones are the neural network (NN) based techniques
\citep{Conv_text, SVM_ANN, Twitter_ngram_ANN, sentiment_deep,
sentiment_NN, fastText}. In \cite{Sparsity}, each element in an input
sequence is first numericalized/vectorized as a vocabulary-sized one-hot
vector with bit ``1" occupying the position corresponding to the index
of that word in the vocabulary. This vector is then fed into a trainable
dense network called the embedding layer. The output of the embedding
layer is another vector of a reduced size. In \cite{Word2vec}, the
embedding layer is integrated into a recurrent NN (RNN) used for
language modeling so that the trained embedding layer can be applied to
more generic language tasks.  Both \cite{Sparsity} and \cite{Word2vec}
conduct dimension reduction at the word level. Hence, they are called
word embedding methods. These methods are limited in modeling
``sequences of words", which is called the sequence-to-vector (seq2vec)
problem, for two reasons. First, word embedding is trained on some
particular dataset using the stochastic gradient descent method, which
could lead to overfitting \citep{GoodEmbedding} easily. Second, the
vector space obtained by word embedding is still too large, it is
desired to convert a sequence of words to an even more compact form. 

Among non-neural-network dimension reduction methods \citep{LSA,
Twitter_ontology, wordnet_chains, sentiment_travel, feature_TC,
TFIGM_TC}, the principal component analysis (PCA) is a popular one. In
\cite{LSA}, sentences are first represented by vocabulary-sized vectors,
where each entry holds the frequency of a particular word in the
vocabulary.  Each sentence vector forms a column in the input data
matrix. Then, the PCA is used to generate a transform matrix for
dimension reduction on each sentence.  Although the PCA has
some nice properties such as maximum information preservation
\citep{PCA_MI} between its input and output under certain constraints, we
will show later that its computational complexity is exceptionally high as
the dataset size becomes large. Furthermore, most non-RNN-based
dimension reduction methods, such as \cite{LSA, feature_TC, TFIGM_TC},
do not consider the positional correlation between elements in a
sequence but adopt the ``bag-of-word" (BoW) representation. The
sequential information is lost in such a dimension reduction procedure. 

To address the above-mentioned shortcomings, a novel technique, called
the tree-structured multi-stage PCA (TMPCA), was proposed in
\cite{TMPCA_ICPR}. The TMPCA method has several interesting properties
as summarized below.
\begin{enumerate}
\item {\bf High efficiency.} Reduce the input data dimension with a
small model size at low computational complexity. 
\item {\bf Low information loss.} Maintain high mutual information
between an input and its dimension-reduced output. 
\item {\bf Sequential preservation.} Preserve the positional
relationship between input elements. 
\item {\bf Unsupervised learning.} Do not demand labeled training data. 
\item {\bf Transparent mathematical properties.} Like PCA, TMPCA is linear 
and orthonormal, which makes the mathematical analysis of the system easier.
\end{enumerate}
These properties are beneficial to classification tasks that demand 
low-dimensional yet highly informative data. It also relaxes the burden of 
data labeling in the training stage. So TMPCA can be used as a preprocessing 
stage for classification problems, a complete classification framework using 
TMPCA is shown in figure below where the training TMPCA does not demand labels:

%%%%%%%%%%%%%%%%%%%%%%%%%%%%%%%%%%%%%%%%%%%%%%%%%%%%%%%%%%%%
\begin{figure*}[htbp]
\centering
\includegraphics[width=0.6\linewidth]{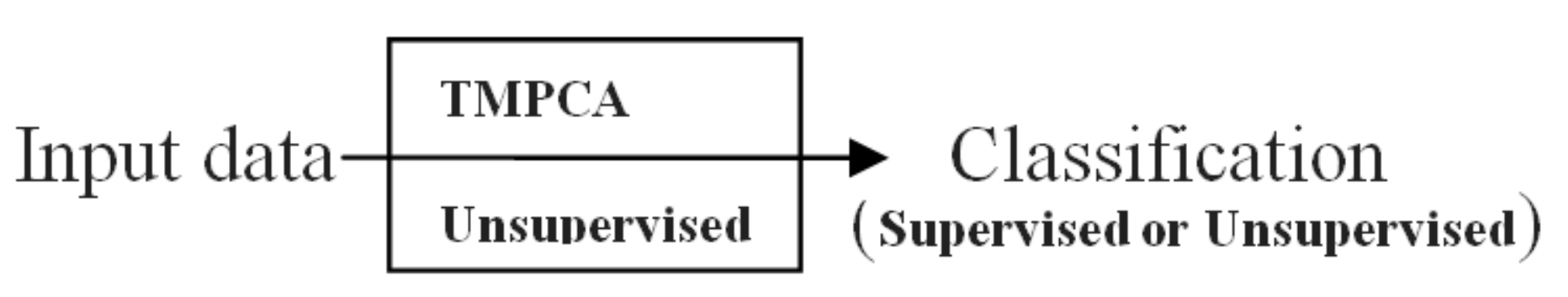} 
\caption{Integration of TMPCA to classification problems.}\label{fig:FrameWork}
\end{figure*}
%%%%%%%%%%%%%%%%%%%%%%%%%%%%%%%%%%%%%%%%%%%%%%%%%%%%%%%%%%%%

In this work, we present the TMPCA method and apply it to several text
classification problems such as spam email detection, sentiment
analysis, news topic identification, etc. This work is an extended
version of \cite{TMPCA_ICPR}. As compared with \cite{TMPCA_ICPR}, most
material in Sec. \ref{sec:TMPCA} and Sec. \ref{sec:exp} is new.  We
present more thorough mathematical treatment in Sec. \ref{sec:TMPCA} by
deriving the function of the TMPCA method and analyzing its properties.
Specifically, the information preserving property of the TMPCA method is
demonstrated by examining the mutual information between its input and
output. Also, we provide more extensive experimental results on large
text classification datasets to substantiate our claims in Sec.
\ref{sec:exp}. 

The rest of this paper is organized as follows. Research on text
classification problems is reviewed in Sec.  \ref{sec:TC}. The TMPCA
method and its properties are presented in Sec.  \ref{sec:TMPCA}.
Experimental results are given in Sec.  \ref{sec:exp}, where we compare
the performance of the TMPCA method and that of state-of-the-art
NN-based methods on text classification, including fastText
\citep{fastText} and the convolutional-neural-network (CNN) based method
\citep{Conv_text}. Finally, concluding remarks are drawn in Sec.
\ref{sec:con}. 

\section{Review of Previous Work on Text Classification}\label{sec:TC}

Text classification has been an active research topic for two decades.
Its applications such as spam email detection, age/gender identification
and sentiment analysis are omnipresent in our daily lives. Traditional
text classification solutions are mostly linear and based on the BoW
representation. One example is the naive Bayes (NB) method \citep{NB},
where the predicted class is the one that maximizes the posterior
probability of the class given an input text. The NB method offers
reasonable performance on easy text classification tasks, where the
dataset size is small.  However, when the dataset size becomes larger,
the conditional independence assumption used in likelihood calculation
required by the NB method limits its applicability to complicated text
classification tasks. 

Other methods such as the support vector machine (SVM) \citep{SVM_ANN,
sentiment_travel, TC_SVM} fit the decision boundary in a hand-crafted
feature space of input texts. Finding representative features of input
texts is actually a dimension reduction problem.  Commonly used features
include the frequency that a word occurs in a document, the
inverse-document-frequency (IDF), the information gain \citep{feature_TC,
TFIGM_TC, IDF, TC_COMP}, etc. Most SVM models exploit BoW features, and
they do not consider the position information of words in sentences. 

The word position in a sequence can be better handled by the CNN
solutions since they process the input data in sequential order. One
example is the character level CNN (char-CNN) as proposed in
\cite{Conv_text}. It represents an input character sequence as a
two-dimensional data matrix with the sequence of characters along one
dimension and the one-hot embedded characters along the other one.
Any character exceeding the maximum allowable sequence length is
truncated.  The char-CNN has 6 convolutional (conv) layers and 3
fully-connected (dense) layers.  In the conv layer, one dimensional
convolution is carried out on each entry of the embedding vector. 

RNNs offer another NN-based solution for text classification
\citep{overview_TC, sentiment_NN}. An RNN generates a compact yet rich
representation of the input sequence and stores it in form of hidden
states of the memory cell. It is the basic computing unit in an RNN.
There are two popular cell designs: the long short-term memory (LSTM)
\citep{LSTM} and the gate recurrent unit (GRU) \citep{GRU}.  Each cell
takes each element from a sequence sequentially as its input, computes
an intermediate value, and updates it dynamically. Such a value is called
the constant error carousal (CEC) in the LSTM and simply a hidden state
in the GRU. Multiple cells are connected to form a complete RNN.  The
intermediate value from each cell forms a vector called the hidden
state.  It was observed in \cite{Time} that, if a hidden state is
properly trained, it can represent the desired text patterns compactly,
and similar semantic word level features can be grouped into clusters.
This property was further analyzed in \cite{DBRNN_ELSTM}. Generally
speaking, for a well designed representational vector (i.e. the hidden
state), the computing unit (or the memory cell) can exploit the
word-level dependency to facilitate the final classification task. 

\begin{figure*}[htbp]
\centering
\includegraphics[width=0.5\linewidth]{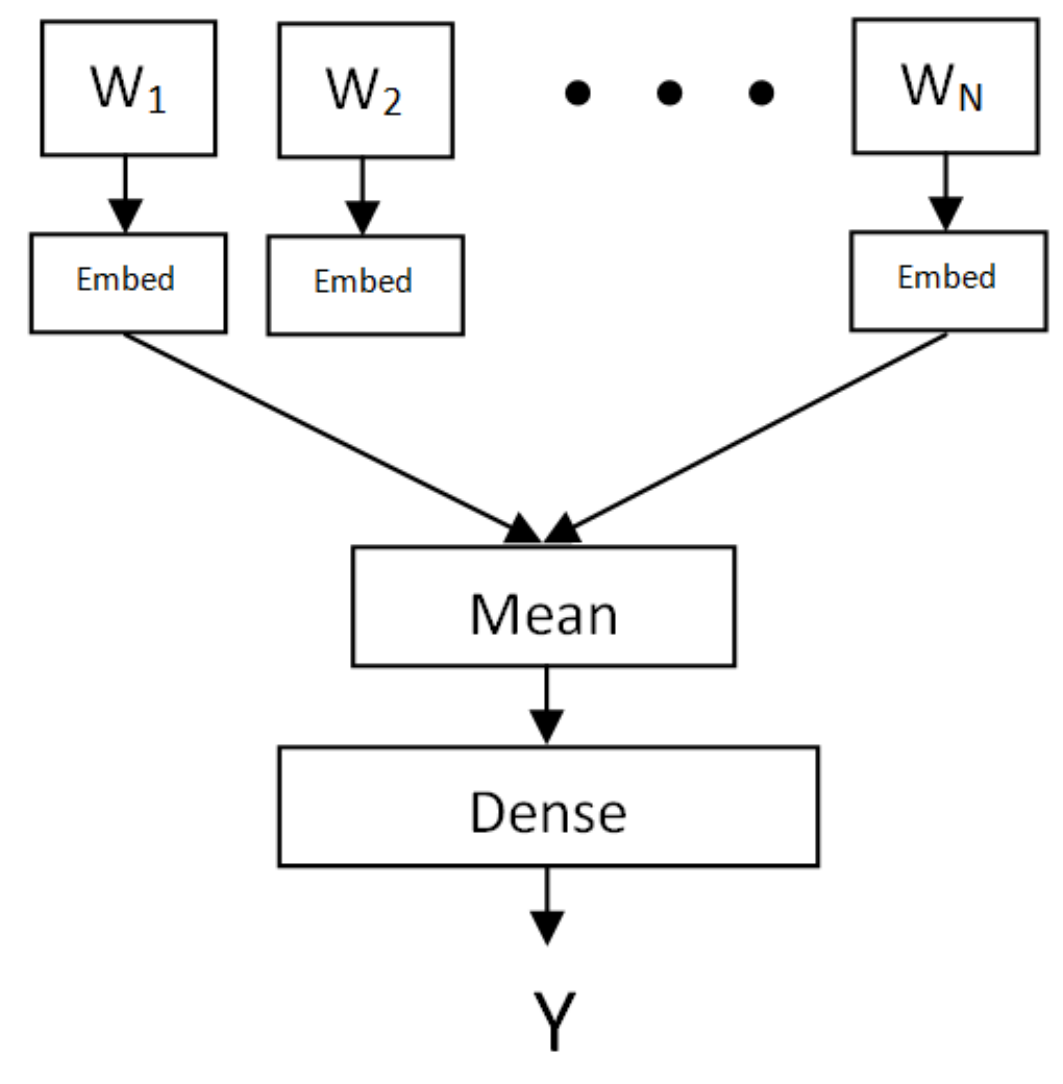} 
\caption{Illustration of the fastText model.}\label{fig:fastText}
\end{figure*}

Another NN-based model is the fastText \citep{fastText}. As shown in Fig.
\ref{fig:fastText}, it is a multi-layer perceptron composed by a
trainable embedding layer, a hidden mean layer and a softmax dense
layer. The hidden vector is generated by averaging the embedded word,
which makes the fastText a BoW model.  The fastText offers a very fast
solution to text classification. It typically takes less than a minute
in training a large data corpus with millions of samples. It gives the
state-of-the-art performance. We would like to use it as the primary
benchmarking algorithm in Sec.  \ref{sec:exp}.  All NN-based text
classification solutions demand labeled data in training.  We will
present the TMPCA method, which does not need labeled training data, in
the next section. 

%%%%%%%%%%%%%%%%%%%%%%%%%%%%%%%%%%%%%%%%%%%%%%%%%%%%%%%%%%%%
%%%%%%%%%%%%%%%%%%%%%%%%%%%%%%%%%%%%%%%%%%%%%%%%%%%%%%%%%%%%

\section{Proposed TMPCA Method}\label{sec:TMPCA}

\begin{figure*}[htbp]
\centering
\includegraphics[width=\linewidth]{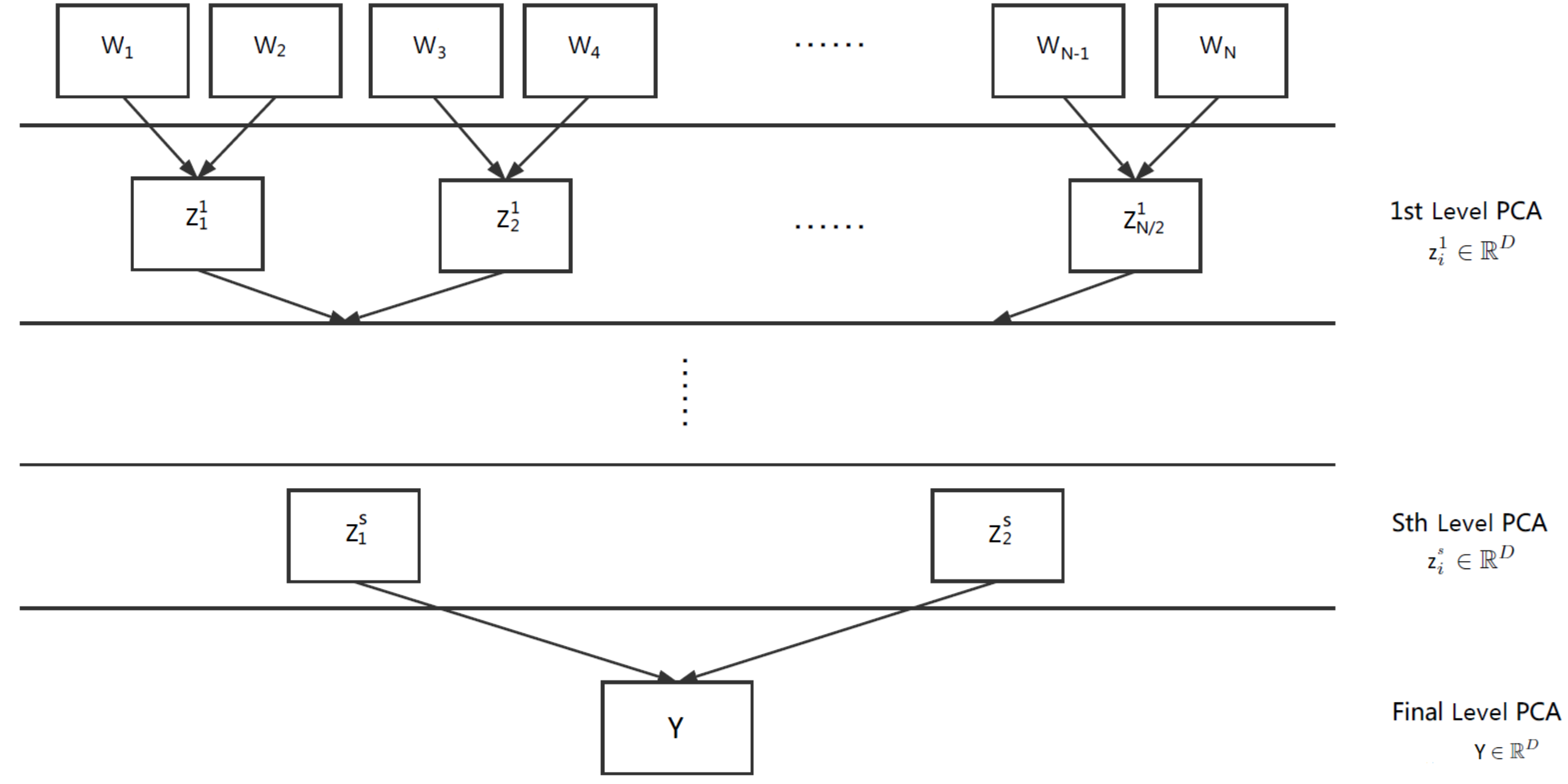} 
\caption{The Block diagram of the TMPCA method.}\label{fig:TMPCA}
\end{figure*}

In essence, TMPCA is a tree-structured multi-stage PCA method whose
input at every stage is two adjacent elements in an input sequence
without overlap.  The reason for every two elements rather than other
number of elements is due to the computational efficiency of such an
arrangement. This will be elaborated in Sec. \ref{sec:TMPCA_complx}. The
block diagram of TMPCA with a single sequence $\{w_1, ..., w_N\}$ as its
input is illustrated in Fig.  \ref{fig:TMPCA}. The input sequence length
is $N$, where $N$ is assumed to be a number of the power of 2 for ease
of discussion below. We will relax such a constraint for practical
implementation in Sec. \ref{sec:exp}. We use $z^s_j$ to denote the
$j$th element in the output sequence of stage $s$ (or equivalently, the
input sequence of stage $s+1$ if such a stage exists).  It is obvious that
the final output $Y$ is also $z^{\text{log}_2N}_1$. 

%%%%%%%%%%%%%%%%%%%%%%%%%%%%%%%%%%%%%%%%%%%%%%%%%%%%%%%%%%%%
%%%%%%%%%%%%%%%%%%%%%%%%%%%%%%%%%%%%%%%%%%%%%%%%%%%%%%%%%%%%

\subsection{Training of TMPCA}\label{sec:TMPCA_train}

To illustrate how TMPCA is trained, we use an example of a training
dataset with two sequences, each of which has four numericalized
elements. Each element is a column vector of size $D$, denoted as
$w^i_j$, where $i$ indicates the corresponding sequence and $j$ is the
position of the element in that sequence.  At each stage of the TMPCA
tree, every two adjacent elements without overlap are concatenated to
form one vector of dimension $2D$. It serves as a sample for PCA
training at that stage. Thus, the training data matrix for PCA at the
first stage can be written as
$$
\begin{bmatrix} 
(w^1_1)^T& (w^1_2)^T \\ 
(w^1_3)^T& (w^1_4)^T \\
(w^2_1)^T& (w^2_2)^T \\ 
(w^2_3)^T& (w^2_4)^T 
\end{bmatrix}.
$$ 

The trained PCA transform matrix at stage $s$ is denoted as $U^s$. It 
reduces the dimension of the input vector from $2D$ to $D$. That is, 
$U^s \in \mathbb{R}^{D \times2D}$. The training matrix at the first 
stage is then transformed by $U^1$ to
$$
\begin{bmatrix}
(z^1_1)^T \\ (z^1_2)^T \\
(z^1_3)^T \\ (z^1_4) \end{bmatrix}, \quad
z^1_1 = U^1(\begin{bmatrix} w^1_1 \\ w^1_2\end{bmatrix}), \quad
z^1_2 = U^1(\begin{bmatrix}w^1_3 \\ w^1_4\end{bmatrix}), \quad
z^1_3 = U^1(\begin{bmatrix} w^2_1 \\ w^2_2\end{bmatrix}), \quad
z^1_4 = U^1(\begin{bmatrix}w^2_3 \\ w^2_4\end{bmatrix}),
$$ 
After that, we rearrange the elements on the transformed training 
matrix to form
$$
\begin{bmatrix} 
(z^1_1)^T& (z^1_2)^T \\ 
(z^1_3)^T& (z^1_4)^T 
\end{bmatrix}.
$$ 

It serves as the training matrix for the PCA at the second stage. We
repeat the training data matrix formation, the PCA kernel determination
and the PCA transform steps recursively at each stage until the length
of the training samples becomes 1.  It is apparent that, after one-stage
TMPCA, the sample length is halved while the element vector size keeps
the same as $D$.  The dimension evolution from the initial input data to
the ultimate transformed data is shown in Table \ref{tab:dim_change}.
Once the TMPCA is trained, we can use it to transform test data by
following the same steps except that we do not need to compute the PCA
transform kernels at each stage. 

%%%%%%%%%%%%%%%%%%%%%%%%%%%%%%%%%%%%%%%%%%%%%%%%%%%%%%%%%%%%
\begin{table}[htbp]
\caption{Dimension evolution from the input to the output in the 
TMPCA method.}\label{tab:dim_change}
\begin{center}
\begin{tabular}{l c c}
\hline
& Sequence length   & Element vector size \\
\hline
Input sequence & N & D\\ \hline
Output sequence& 1& D \\ \hline
\end{tabular}
\end{center}
\end{table}
%%%%%%%%%%%%%%%%%%%%%%%%%%%%%%%%%%%%%%%%%%%%%%%%%%%%%%%%%%%%

\subsection{Computational Complexity}\label{sec:TMPCA_complx}

We analyze the time complexity of TMPCA training in this section.
Consider a training dataset of $M$ samples, where each sample is of
length $N$ with element vectors of dimension $D$. To determine the PCA
model for this training matrix of dimension $\mathbb{R}^{M \times ND}$,
it requires $\text{O}(MN^2D^2)$ to compute the covariance matrix, and
$\text{O}(N^3D^3)$ to compute the eigenvalues of the covariance matrix.
Thus, the complexity of PCA can be written as
\begin{equation}\label{eq:PCA_complx}
\text{O}(f_{\text{PCA}}) = \text{O}\Big(N^3 D^3 + M N^2 D^2 \Big).
\end{equation} 
The above equation can be simplified by comparing the value of $M$ with
$ND$. We do not pursue along this direction furthermore since it is problem
dependent. 

Suppose that we concatenate non-overlapping $P$ adjacent elements at each
stage of TMPCA. The dimension of the training matrix at stage $s$ is
$M\frac{N}{P^s}\times PD$.  Then, the total computational complexity of
TMPCA can be written as
\begin{align} \label{eq:TMPCA_complx}
\text{O}(f_{\text{TMPCA}}) 
& = \text{O}\Big(\sum^{\log_PN}_{s=1}\big((PD)^3 + M\frac{N}{P^s}(PD)^2 \big)\Big),\nonumber\\
& = \text{O}\Big((P^3 \text{log}_PN) D^3 + M\frac{P^2}{P-1}(N-1)D^2\Big).
\end{align}

The complexity of TMPCA is an increasing function in $P$.  This can be
verified by non-negativity of its derivative with respect to $P$.  Thus,
the worst case is $P=N$, which is simply the traditional PCA applied to
the entire samples in a single stage.  When $P=2$, the TMPCA achieves
its optimal efficiency. Its complexity is
\begin{align}\label{eq:TMPCA_complx_2}
\text{O}(f_{\text{TMPCA}}) 
& = \text{O}\Big(8 (\text{log}_2N) D^3 + 4M(N-1)D^2\Big), \nonumber\\
& = \text{O}\Big(2 (\text{log}_2N) D^3 + M(N-1)D^2\Big).
\end{align}
By comparing Eqs. (\ref{eq:TMPCA_complx_2}) and (\ref{eq:PCA_complx}),
we see that the time complexity of the traditional PCA grows at least 
quadratically with sentence length $N$ (since $P=N$) while that of TMPCA
grows at most linearly with $N$. 

\subsection{System Function}\label{sec:TMPCA_sys}

To analyze the properties of TMPCA, we derive its system function in
closed form in this section. In particular, we will show that, similar
to PCA, TMPCA is a linear transform and its transformation matrix has
orthonormal rows.  For the rest of this paper, we assume that the length
of the input sequence is $N$ $N=2^L$, where $L$ is the total stage
number of TMPCA. The input is mean-removed so that its mean is $\bm{0}$. 

We denote the element of the input sequence by $w_j$, where $w_j \in
\mathbb{R}^D$ and $j \in \{1,...,N\}$.  Then, the input sequence $X$ to
TMPCA is a column vector in form of
\begin{equation}
X^T = [w^T_1, \cdots , w^T_N]. 
\end{equation}
We decompose PCA transform matrices, $U^s$, at stage $s$ into two
equal-sized block matrices as
\begin{equation}
U^s = [U^s_1, U^s_2],
\end{equation}
where $U^s_j \in \mathbb{R}^{D\times D}$, and where $j \in \{1,2\}$. 
The output of TMPCA is $Y \in \mathbb{R}^D$

With notations inherited from Sections \ref{sec:TMPCA_train} and
\ref{sec:TMPCA_complx}, we can derive the closed-form expression of
TMPCA by induction (see Appendix A). That is, we have
\begin{align}\label{eq:TMPCA_func}
Y &= UX, \\
U &= [U_1, ..., U_N],\\
U_j &= \prod^{L}_{s=1} U^s_{f_{j,s}}, \forall j\in\{1,...,N\}\\
f_{j,s} &= \text{b}_L(j-1)_s+1, \forall j,s.
\end{align}
where $\text{b}_L(x)_s$ is the $s$th digit of $L$-binarized form of
$x$. TMPCA is a linear transform as shown in Eq. \ref{eq:TMPCA_func}. 
Also, since there always exist real valued eigenvectors to form the PCA 
transform matrix, $U$, $U_j$ and $\{U^s_j\}^2_{j=1}$ are all real valued 
matrices.

To show that $U$ has orthonormal rows, we first examine the
properties of matrix $K = [U_1, U_2]$. By setting
$$
A = \prod^{L}_{s=2} U^s_{f_{1,s}} = \prod^{L}_{s=2} U^s_{f_{2,s}},
$$
we obtain $K = [A U^1_1, A U^1_2]$. Since matrix $[U^1_1, U^1_2]$ is a PCA
transform matrix, it has orthonormal rows. Denote $<\cdot>_{ij}$ as the 
inner product between the $i$th row and $j$th row of matrix $\cdot$, we 
conclude that the $<K>_{ij} = <A>_{ij}$ using the following property. 
\begin{lemma}\label{lem:orthonormal}
Given $K=[A B_1, A B_2]$, where $[B_1, B_2]$ has orthonormal rows, then 
$<K>_{ij} = <A>_{ij}$. 
\end{lemma}
We then let
\begin{equation}\label{eq:TMPCA_K}
K^s_m = [A^s_mU^s_1, A^s_mU^s_2], 
\end{equation}
where $s \in \{1,...,L\}$ indicates the stage, and $m \in \{1,...,
\frac{N}{2^s}\}$, and 
\begin{equation}\label{eq:TMPCA_A}
A^s_m = \prod^{L}_{k=s+1} U^k_{f_{m,k-s}}, \mbox{ and } A^L_1 = \bm{I},
\end{equation}
where $\bm{I}$ is the identity matrix. At stage 1, $K^1_m = [U_{2m-1}, U_{2m}]$, so $U = [K^1_1, ..., K^1_{N/2}]$.
Since $<U>_{ij} = \sum^{N/2}_{m=1}<K^1_m>_{ij}$, according to 
\ref{lem:orthonormal}, Eqs. (\ref{eq:TMPCA_K})-(\ref{eq:TMPCA_A}) we have 
\begin{align}\label{eq:TMPCA_orthonormality}
<U>_{ij} & = \sum^{N/2}_{m=1}<A^1_m>_{ij}, \nonumber \\
		 & = \sum^{N/4}_{m=1}<K^2_m>_{ij} = \sum^{N/4}_{m=1}<A^2_m>_{ij},
             	\nonumber \\
         & \hdots \nonumber\\
         & = <K^L_1>_{ij} = <[U^L_1, U^L_2]>_{ij},
\end{align}
Thus, $U$ has orthonormal rows. 

\subsection{Information Preservation Property}\label{sec:TMPCA_info}

Besides its low computation complexity, linearity and orthonormality,
TMPCA can preserve the information of its input effectively so as to
facilitate the following classification process. To show this point, 
we investigate the mutual information \citep{PCA_MI, JMIM} between 
the input and the output of TMPCA.

Here, the input to the TMPCA system is modeled as
\begin{equation}
X = G + n,
\end{equation}
where $G$ and $n$ are used to model the ground truth semantic signal and
the noise component in the input, respectively. In other words, $G$
carries the essential information for the text classification task while
$n$ is irrelevant to (or weakly correlated) the task.  We are interested
in finding the mutual information between output $Y$ and ground truth
$G$. 

By following the framework in \cite{PCA_MI}, we make the following assumptions:
\begin{enumerate}
\item $Y \sim \mathbb{N}(\bar{y}, \bm{V})$;
\item $n \sim \mathbb{N}(\bm{0}, \bm{B})$, where $\bm{B} = \sigma^2\bm{I}$;
\item $n$ is uncorrelated with $G$.
\end{enumerate}
In above, $\mathbb{N}$ denotes the multivariant Gaussian density function. 
Then, the mutual information between $Y$ and $G$ can be computed as
\begin{align}
I(Y,G) = &\text{E}_{Y,G} \left( \text{ln}\frac{P(Y|G)}{P(Y)} \right),\nonumber\\
   = &\text{E}_{Y,G} \left( \text{ln}\frac{\mathbb{N}(Ug,U\bm{B}U^T)}
   {\mathbb{N}(\bar{y},\bm{V})} \right), \nonumber\\
   = &\frac{1}{2}\text{ln}\frac{|V|}{|U\bm{B}U^T|}
   - \frac{1}{2}\text{E}_{Y,G}\left[ (y-Ug)^T(UBU^T)^{-1}(y-Ug)\right]  \nonumber\\
   &+ \frac{1}{2}\text{E}_{Y,G}\left[ (y-\bar{y})^TV^{-1}(y-\bar{y})\right],
\end{align}
where $y \in Y$, $g \in G$, and $P(\cdot)$, $|\cdot|$ and $E_X$ denote
the probability density function, the determinant and the expectation of
random variable $X$, respectively. It is straightforward to prove the
following lemma,
\begin{lemma}\label{lem:expectation}
For any random vector $X \in \mathbb{R}^D$ with covariance matrix $K_x$, the
following equality holds
$$
E_X\{(x-\bar{x})^T(K_x)^{-1}(x-\bar{x})\} = D.
$$
\end{lemma}
Then, based on this lemma, we can derive that
\begin{equation}\label{eq:TMPCA_mi}
I(Y,G) = \frac{1}{2}\text{ln}\frac{|V|}{\sigma^{2D}}.
\end{equation}

The above equation can be interpreted below.  Given input signal noise
$\sigma$, the mutual information can be maximized by maximizing the
determinant of the output covariance matrix. Since TMPCA maximizes the
covariance of its output at each stage, TMPCA will deliver an output
with the largest mutual information at the corresponding stage. We will
show experimentally in Section \ref{sec:exp} that the mutual information
of TMPCA is significantly larger than that of the mean operation and
close to that of PCA. 

\section{Experiments}\label{sec:exp}

\subsection{Datasets}\label{sec:exp_data}

We tested the performance of the TMPCA method on twelve datasets of
various text classification tasks as shown in Table \ref{tab:exp_data}.
Four of them are smaller datasets with at most 10,000 training samples.
The other eight are large-scale datasets \citep{Conv_text} with training
samples ranging from 120 thousands to 3.6 millions. 

%%%%%%%%%%%%%%%%%%%%%%%%%%%%%%%%%%%%%%%%%%%%%%%%%%%%%%%%%%%%%
\begin{table*}[htbp]
\centering
\caption{Selected text classification datasets.}\label{tab:exp_data}
\begin{tabular*}{\textwidth}{@{\extracolsep{\fill} }l  c c c c}
\hline
&  \# of Class & Train Samples & Test Samples & \# 
of Tokens  \\ \hline
spam &  2 & 5,574  & 558 & 14,657 \\ \hline
sst & 2 & 8409  & 1803 & 18,519\\\hline
semeval & 2 & 5098  & 2034 & 25,167 \\\hline
imdb & 2 & 10162  & 500 & 20,892 \\\hline
agnews & 4 & 120,000 & 7,600 & 188,111 \\\hline
sogou & 5 & 450,000 & 60,000 & 800,057 \\\hline
dbpedia & 14 & 560,000 & 70,000 &  1,215,996  \\\hline
yelpp & 2 & 560,000 & 38,000& 1,446,643 \\\hline
yelpf & 5 & 650,000 & 50,000 & 1,622,077 \\\hline
yahoo & 10 & 1,400,000 & 60,000 & 4,702,763 \\\hline
amzp & 2 & 3,600,000 & 400,000 & 4,955,322 \\\hline
amzf & 5 & 3,000,000 & 650,000 & 4,379,154 \\\hline
\end{tabular*}
\end{table*}
%%%%%%%%%%%%%%%%%%%%%%%%%%%%%%%%%%%%%%%%%%%%%%%%%%%%%%%%%%%%%

These datasets are briefly introduced below.
\begin{enumerate}
\item {\bf SMS Spam (spam)} \citep{data_SMS_SPAM}. It is a dataset
collected for mobile Spam email detection. It has two target classes:
``Spam" and ``Ham". 
\item {\bf Stanford Sentiment Treebank (sst)} \citep{data_SST}. It is a
dataset for sentiment analysis. The labels are generated using the
Stanford CoreNLP toolkit \citep{SST_NLP_tool}. The sentences labeled as
very negative or negative are grouped into one negative class. Sentences
labeled as very positive or positive are grouped into one positive
class.  We keep only positive and negative sentences for training and
testing. 
\item {\bf Semantic evaluation 2013 (semeval)} \citep{data_SEMEVAL}. It
is a dataset for sentiment analysis. We focus on Sentiment task-A with
positive/negative two target classes. Sentences labeled as ``neutral"
are removed. 
\item {\bf Cornell Movie review (imdb)} \citep{data_IMDB}. It is a
dataset for sentiment analysis for movie reviews. It contains a
collection of movie review documents with their sentiment polarity
(i.e., positive or negative). 
\item {\bf AG's news (agnews)} \citep{data_convtext}. It is a dataset for
news categorization. Each sample contains the news title and
description. We combine the title and description into one sentence by
inserting a colon in between. 
\item {\bf Sougou news (sogou)} \citep{data_convtext}. It is a Chinese
news categorization dataset. Its corpus uses a phonetic romanization of
Chinese. 
\item {\bf DBPedia (dbpedia)} \citep{data_convtext}. It is an ontology
categorization dataset with its samples extracted from the Wikipedia.
Each training sample is a combination of its title and abstract. 
\item {\bf Yelp reviews (yelpp and yelpf)} \citep{data_convtext}. They
are sentiment analysis datasets. The Yelp review full (yelpf) has
target classes ranging from one to five stars. The one star is
the worst while five stars the best. The Yelp review polarity (yelpp)
has positive/negative polarity labels by treating stars 1 and 2 as
negative, stars 4 and 5 positive and omitting star 3 in the polarity
dataset. 
\item {\bf Yahoo! answers (yahoo)} \citep{data_convtext}. It is a topic
classification dataset for Yahoo's question and answering corpus. 
\item {\bf Amazon reviews (amzp and amzf)} \citep{data_convtext}. These
two datasets are similar to Yelp reviews but of much larger sizes. They
are about Amazon product reviews. 
\end{enumerate}

\subsection{Experimental Setup}\label{sec:exp_setup}

We compare the performance of the following three methods on the four
small datasets:
\begin{enumerate}
\item TMPCA-preprocessed data followed by the dense network (TMPCA+Dense);
\item fastText;
\item PCA-preprocessed data followed by the dense network (PCA+Dense).
\end{enumerate}

For the eight larger datasets, we compare the performance of six methods. 
They are:
\begin{enumerate}
\item TMPCA-preprocessed data followed by the dense network (TMPCA+Dense);
\item fastText;
\item PCA-preprocessed data followed by the dense network (PCA+Dense);
\item char-CNN \citep{Conv_text};
\item LSTM (an RNN based method) \citep{Conv_text};
\item BoW \citep{Conv_text}.
\end{enumerate}

Besides training time, classification accuracy and F1 macro score, we compute the
mutual information between the input and the output of the TMPCA method,
the mean operation (used by fastText for hidden vector computation) and
the PCA method, respectively. Note that the mean operation can be
expressed as a linear transform in form of
\begin{equation}\label{eq:mean}
Y = \frac{1}{N}[\bm{I},...,\bm{I}] X,
\end{equation}
where $\bm{I} \in \mathbb{R}^{D\times D}$ is the identity matrix 
and the mean transform matrix has $N$ $\bm{I}$'s. 
The mutual information between the input and the output of the mean 
operation can be calculated as
\begin{equation}\label{eq:mean_mi}
I(Y,G) = \frac{1}{2}\text{ln}\frac{|V|N^D}{\sigma^{2D}}.
\end{equation}
For fixed noise variance $\sigma^{2}$, we can compare the mutual
information of the input and the output of different operations
by comparing their associated $|V|$, $|V|N^D$.

To illustrate the information preservation property of TMPCA across
multiple stages, we compute the output energy, which is the sum of
squared elements in a vector/tensor, as a percentage of its input
energy, and see how the energy values decrease as the stage number
becomes bigger. Such investigation is meaningful since the energy
indicates signal's variance in a TMPCA system. The variance is a good
indicator of information richness. The energy percentage is an indicator
of the amount of input information that is preserved after one TMPCA
stage. We compute the total energy of multiple sentences by adding them
together. 

To numericalize the input data, we first remove the stop words from
sentences according to the stop-word list, tokenize sentences and, then,
stem tokens using the python natural language toolkit (NLTK).
Afterwards, we use the fastText-trained embedding layer to embed the
tokens into vectors of size 10. The tokens are then concatenated to form
a single long vector. 

In TMPCA, to ensure that the input sequence is of the same length and
equal to a power of 2, we assign a fixed input length, $N = 2^L$, to all
sentences of length $N'$. If $N' < N$, we preprocess the input sequence
by padding it to be of length $N$ with a special symbol. If $N' > N$, we
shorten the input sequence by dividing it into $N$ segments and
calculating the mean of numericalized elements in each segment. The new
sequence is then formed by the calculated means.  The reason of dividing
a sequence into segments is to ensure consecutive elements as close as
possible. Then, the segmentation of an input sequence can be conducted
as follows. 
\begin{enumerate}
\item Calculate the least number of elements that each segment
should have: $d = \text{floor}(N'/N)$, where floor denotes flooring
operation. 
\item Then we allocate the remaining $r = N'-dN$ elements by
adding one more element to every other floor$(N/r)$ segments
until there are no more elements left. 
\end{enumerate}
To give an example, to partition the sequence $\{w_1,
\cdots , w_{10}\}$ into four segments, we have 3, 2, 3, 2 elements in
these four segments, respectively. That is, they are: $\{w_1, w_2,
w_3\}$, $\{w_4, w_5\}$, $\{w_6, w_7, w_8\}$, $\{w_9, w_{10}\}$. 

For large-scale datasets, we calculate the training data covariance
matrix for TMPCA incrementally by calculating the covariance matrix on
each smaller non-overlapping chunk of the data and, then, adding the
calculated matrices together. The parameters used in dense network
training are shown in Table \ref{tab:exp_dense}.  For TMPCA and PCA, the
numericalized input data are first preprocessed to a fixed length and,
then, have their means removed. TMPCA, fastText and PCA were trained on
Intel Core i7-5930K CPU. The dense network was trained on the GeForce
GTX TITAN X GPU. TMPCA and PCA were not optimized for multi-threading
whereas fastText was run on 12 threads in parallel. 

%%%%%%%%%%%%%%%%%%%%%%%%%%%%%%%%%%%%%%%%%%%%%%%%%%%%%%%%%%%%%
\begin{table}[htbp]
\caption{Parameters in dense network training.}\label{tab:exp_dense}
\begin{center}
\begin{tabular}{l c }
\hline\rule{0pt}{12pt}
Input size & 10\\ \hline\rule{0pt}{12pt}
Output size & \# of target class\\\hline\rule{0pt}{12pt}
Training steps & 5 epochs \\\hline\rule{0pt}{12pt}
Learning rate & 0.5 \\\hline\rule{0pt}{12pt}
Training optimizer & Adam \citep{ADAM} \\[2pt] \hline
\end{tabular}
\end{center}
\end{table} 
%%%%%%%%%%%%%%%%%%%%%%%%%%%%%%%%%%%%%%%%%%%%%%%%%%%%%%%%%%%%%

\subsection{Results}\label{sec:exp_res}

\subsubsection{Performance Benchmarking with State-of-the-Art Methods}\label{exp_othermodels}

We report the results of using the TMPCA method for feature extraction
and the dense network for decision making in terms of test accuracy, F1 macro score,
training time and number of model parameters for text classification with
respect to the eight large datasets.  Furthermore, we conduct
performance benchmarking between the proposed TMPCA model against
several state-of-the-art models. 

The bigram training data for the dense network are generated by
concatenating the bigram representation of the samples to their
original. For example, for sample of $\{w_1, w_2, w_3\}$, after the
bigram process, it becomes $\{w_1, w_2, w_3, w_1w_2, w_2w_3\}$. The accuracy 
and training time for models other than TMPCA are from their original reports in
\cite{Conv_text} and \cite{fastText}. There are two char-CNN models.  We
report the test accuracy of the better model in Table
\ref{tab:exp_accu_models} and the time and model complexity of the
smaller model in Tables \ref{tab:exp_time_models} and
\ref{tab:exp_size_models}. The time reported for char-CNN and fastText
in Table \ref{tab:exp_time_models} is for one epoch only. We only report 
the F1 macro score of TMPCA+Dense against the fastText since firstly fastText has 
the best performance among the other models and secondly it takes very long time 
to generate the results for other models (see Table \ref{tab:exp_time_models})

It is obvious that the TMPCA+Dense method is much faster.  Besides, it
achieves better or commensurate performance as compared with other
state-of-the-art methods.  In addition, the number of parameters of
TMPCA is also much less than other models as shown in Table
\ref{tab:exp_size_models}. 

%%%%%%%%%%%%%%%%%%%%%%%%%%%%%%%%%%%%%%%%%%%%%%%%%%%%%%%%%%%%%
\begin{table*}[htbp]
\centering
\caption{Performance comparison (accuracy (\%)/F1 macro) of different TC models.}\label{tab:exp_accu_models}
\begin{tabular*}{\textwidth}{@{\extracolsep{\fill} }l c c c c c}
\hline
& BoW & LSTM & char-CNN & fastText & \parbox[c][1.2cm]{2.7cm}{TMPCA+Dense \\(bigram, $N=8$)}\\ \hline
agnews& 88.8 & 86.1 & 87.2 & 91.5/0.921& \textbf{92.1}/\textbf{0.930}\\ \hline 
sogou& 92.9 & 95.2 & 95.1 & 93.9/0.970 & \textbf{97.0}/\textbf{0.982} \\ \hline 
dbpedia& 96.6 & \textbf{98.6} & 98.3 & 98.1/\textbf{0.986} & \textbf{98.6}/0.981 \\ \hline 
yelpp& 92.2 & 94.7 & 94.7 & 93.8/0.950 & \textbf{95.1}/\textbf{0.958} \\ \hline 
yelpf& 58.0 & 58.2 & 62.0 & 60.4/0.578 &\textbf{64.1}/\textbf{0.594} \\ \hline 
yahoo& 68.9 & 70.8  & 71.2 & \textbf{72.0}/\textbf{0.695} & \textbf{72.0}/0.688\\ \hline 
amzp& 90.4 & 93.9 &\textbf{ 94.5} & 91.2/0.934 & 94.2/\textbf{0.962} \\ \hline 
amzf& 54.6 & 59.4 & \textbf{59.5} & 55.8/0.533 & 59.0/\textbf{0.587}  \\ \hline 
\end{tabular*}
\end{table*}
%%%%%%%%%%%%%%%%%%%%%%%%%%%%%%%%%%%%%%%%%%%%%%%%%%%%%%%%%%%%%

%%%%%%%%%%%%%%%%%%%%%%%%%%%%%%%%%%%%%%%%%%%%%%%%%%%%%%%%%%%%%
\begin{table*}[htbp]
\centering
\caption{Comparison of training time for different models.}\label{tab:exp_time_models}
\begin{tabular*}{\textwidth}{@{\extracolsep{\fill} }l c c c}
\hline
& small char-CNN/epoch & fastText/epoch & \parbox[c][1.2cm]{2.7cm}{TMPCA+Dense \\(bigram, $N=8$)}\\ \hline
agnews& 1h& 1s & \textbf{0.025}s\\ \hline 
sogou& - & 7s & \textbf{0.081}s\\ \hline 
dbpedia& 2h & 2s & \textbf{0.101}s\\ \hline 
yelpp& - & 3s & \textbf{0.106}s\\ \hline 
yelpf& - & 4s & \textbf{0.116}s\\ \hline 
yahoo& 8h& 5s &  \textbf{0.229}s\\ \hline 
amzp& 2d & 10s & \textbf{0.633}s\\ \hline 
amzf& 2d & 9s & \textbf{0.481}s\\ \hline 
\end{tabular*}
\end{table*}
%%%%%%%%%%%%%%%%%%%%%%%%%%%%%%%%%%%%%%%%%%%%%%%%%%%%%%%%%%%%%

%%%%%%%%%%%%%%%%%%%%%%%%%%%%%%%%%%%%%%%%%%%%%%%%%%%%%%%%%%%%%
\begin{table*}[htbp]
\centering
\caption{Comparison of model parameter numbers in different models.}\label{tab:exp_size_models}
\begin{tabular*}{\textwidth}{@{\extracolsep{\fill} }l c c c}
\hline
& small char-CNN/epoch & fastText/epoch & \parbox[c][1.2cm]{2.7cm}{TMPCA+Dense \\(bigram, $N=8$)}\\ \hline
agnews& \multirow{8}{*}{2.7e+06}  & 1.9e+06 & \multirow{8}{*}{600} \\ 
sogou&  & 8e+06 & \\ 
dbpedia& & 1.2e+07 &\\ 
yelpp& & 1.4e+07 & \\
yelpf&  & 1.6e+07  & \\ 
yahoo& & 4.7e+07 &  \\ 
amzp & & 5e+07  & \\
amzf&  & 4.4e+07 & \\ 
\end{tabular*}
\end{table*}
%%%%%%%%%%%%%%%%%%%%%%%%%%%%%%%%%%%%%%%%%%%%%%%%%%%%%%%%%%%%%

\subsubsection{Comparison between TMPCA and PCA}

We compare the performance between TMPCA+Dense and PCA+Dense to shed
light on the property of TMPCA. Their input are unigram data in each
original dataset. We compare their training time in Table
\ref{tab:exp_time}. It clearly shows the advantage of TMPCA in terms of
computational efficiency. TMPCA takes less than one second for training
in most datasets. As the length of the input sequence is longer, the
training time of TMPCA grows linearly. In contrast, it grows much faster
in the PCA case. 

\begin{figure}[htbp]
\hspace*{-0.1in}
\centering
\subfloat[]{\label{fig:accu_spam} \includegraphics[width 
= 0.33\linewidth]{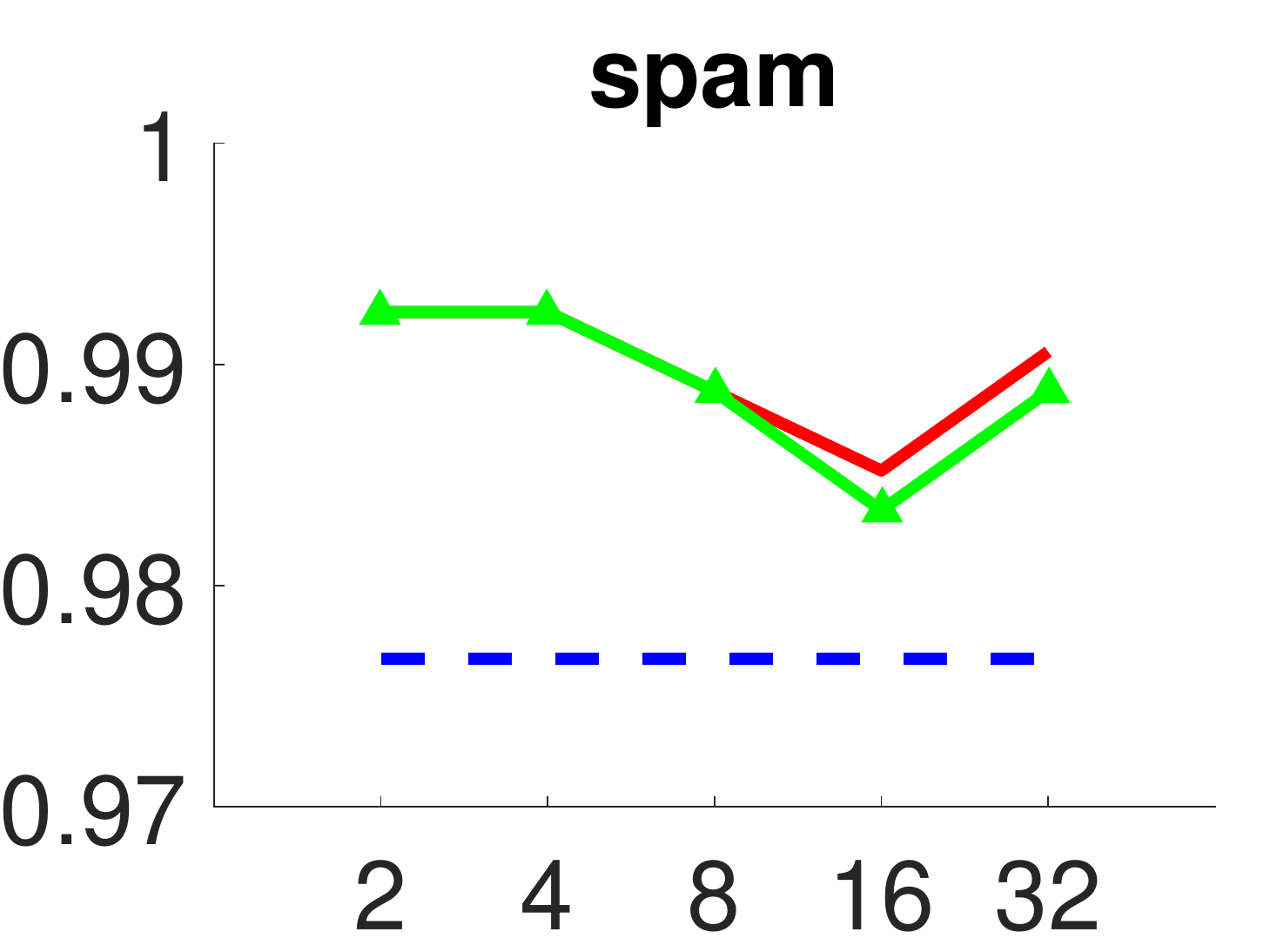}}
\centering
\subfloat[]{\label{fig:accu_sst} \includegraphics[width 
= 0.33\linewidth]{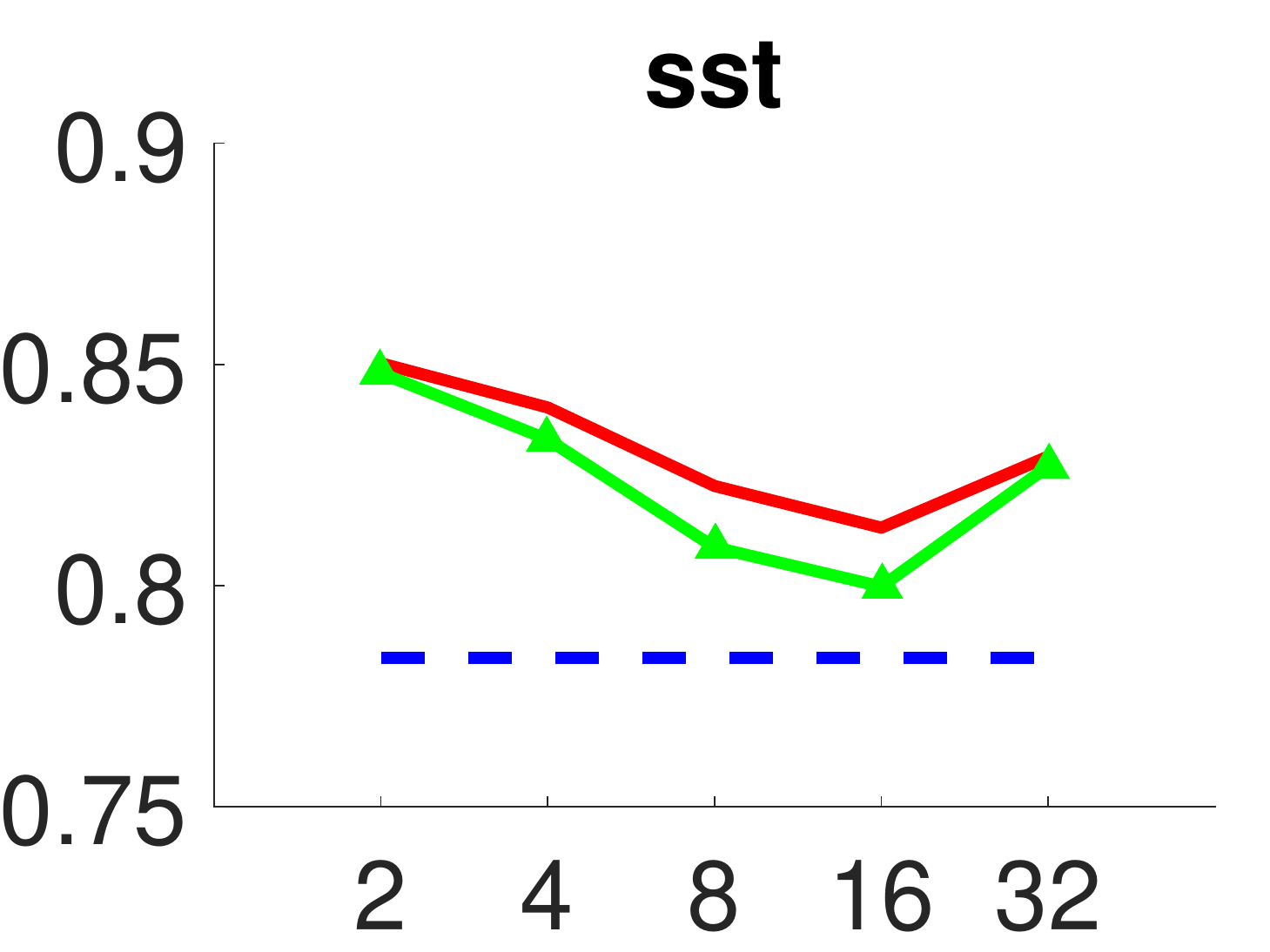}}
\centering
\subfloat[]{\label{fig:accu_semeval} \includegraphics[width 
= 0.33\linewidth]{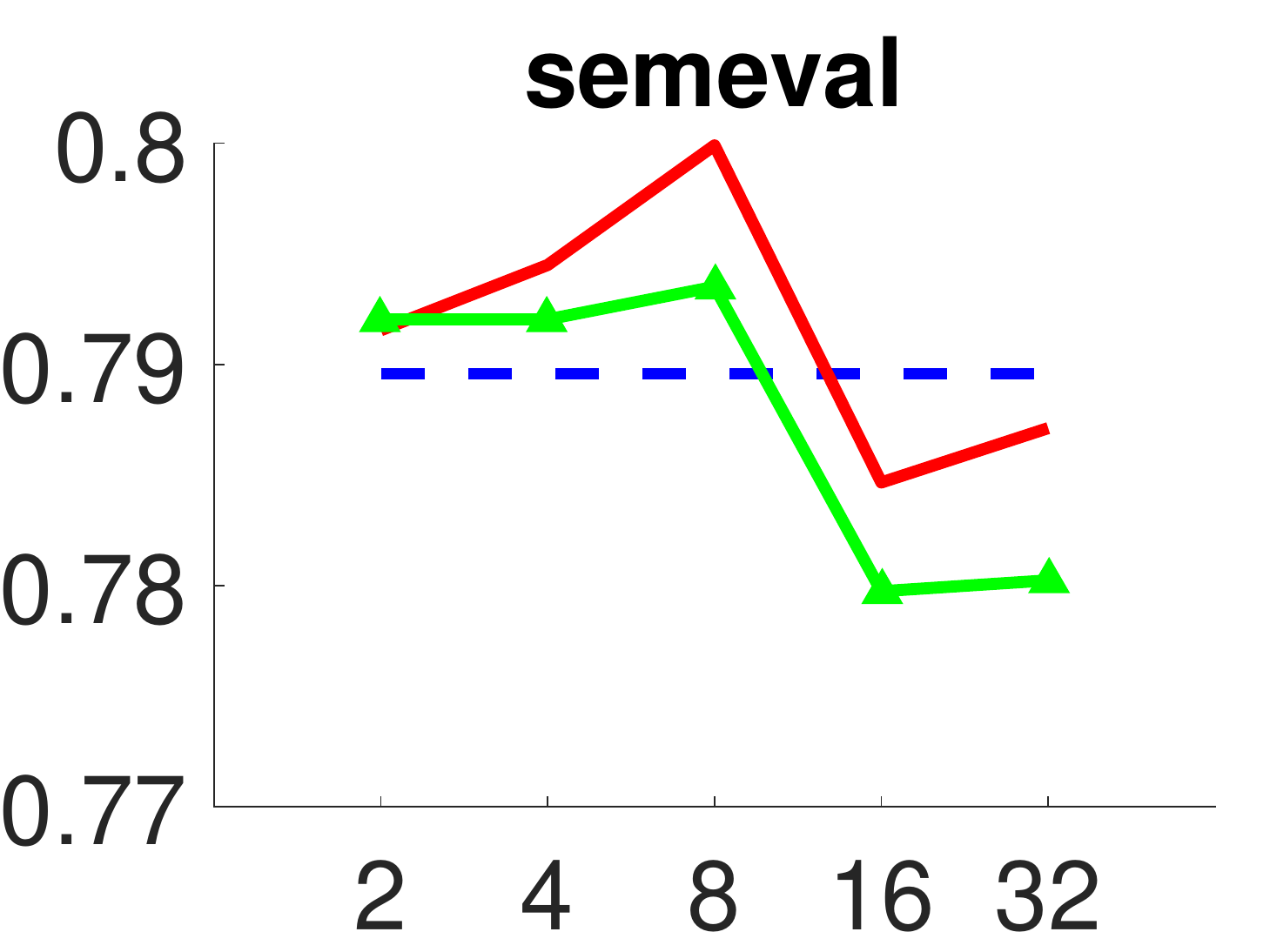}} \\
\centering
\subfloat[]{\label{fig:accu_imdb} \includegraphics[width 
= 0.33\linewidth]{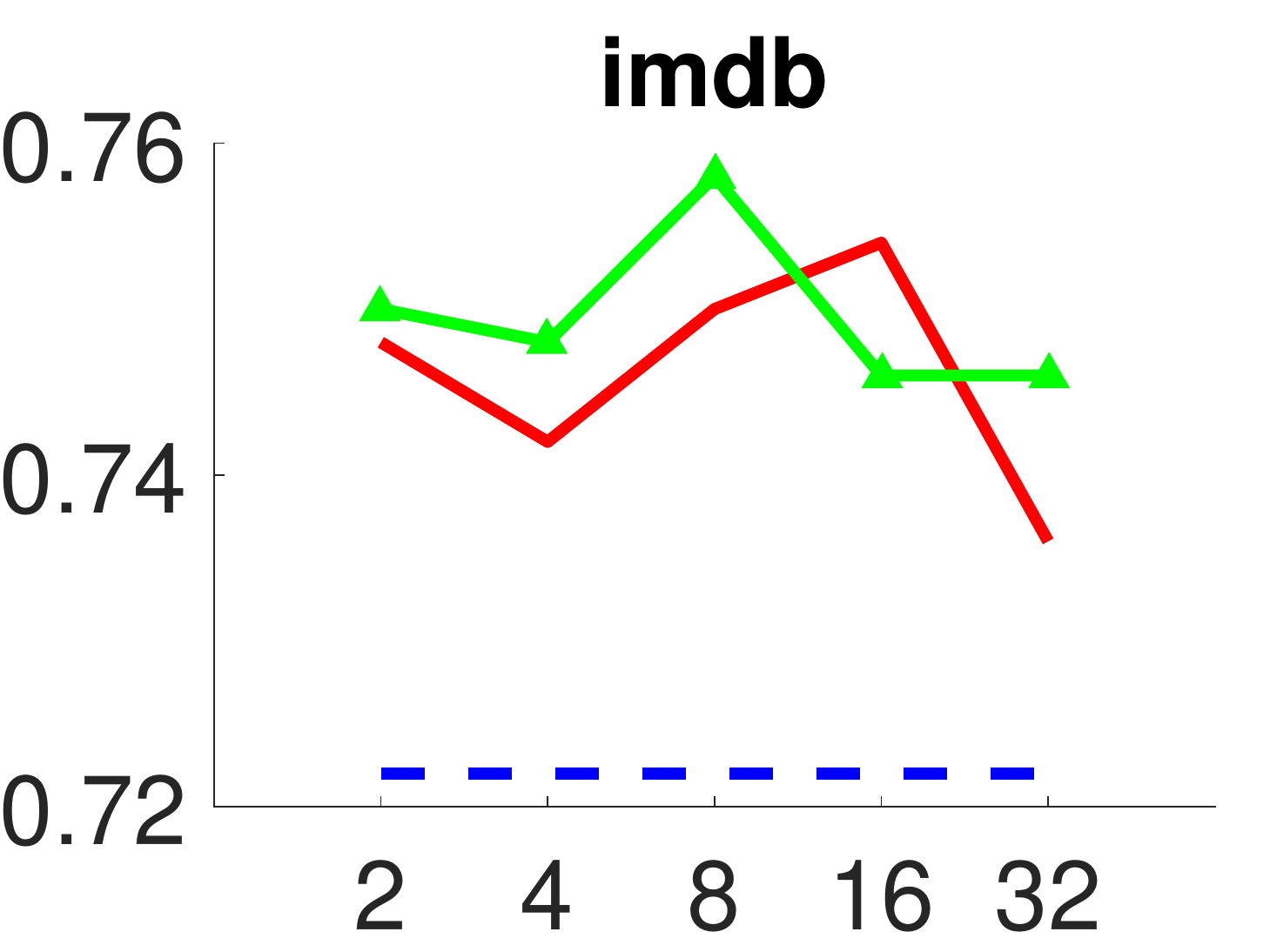}}
\centering
\subfloat[]{\label{fig:accu_agnews} \includegraphics[width 
= 0.33\linewidth]{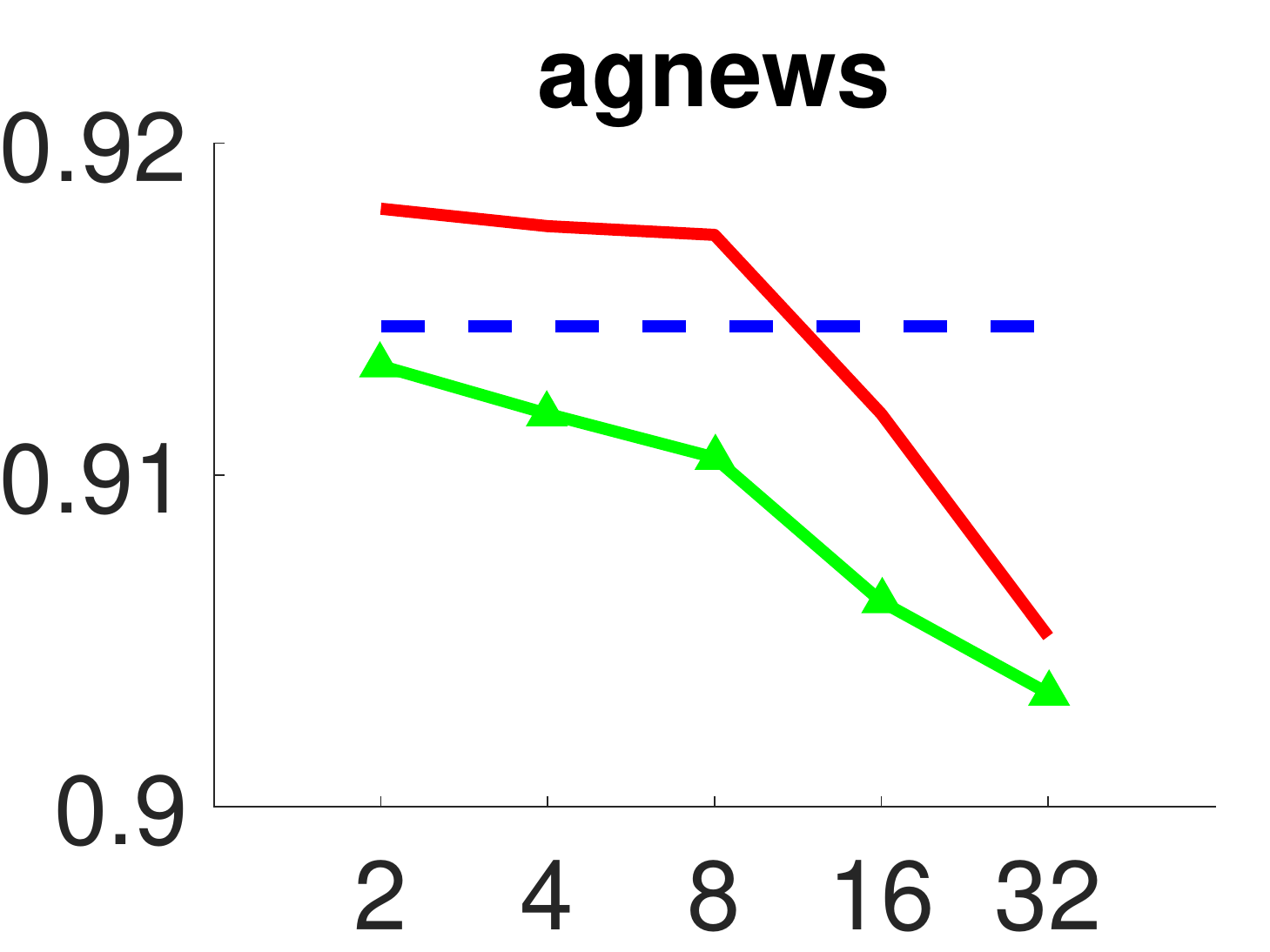}}
\centering
\subfloat[]{\label{fig:accu_sogou} \includegraphics[width 
= 0.33\linewidth]{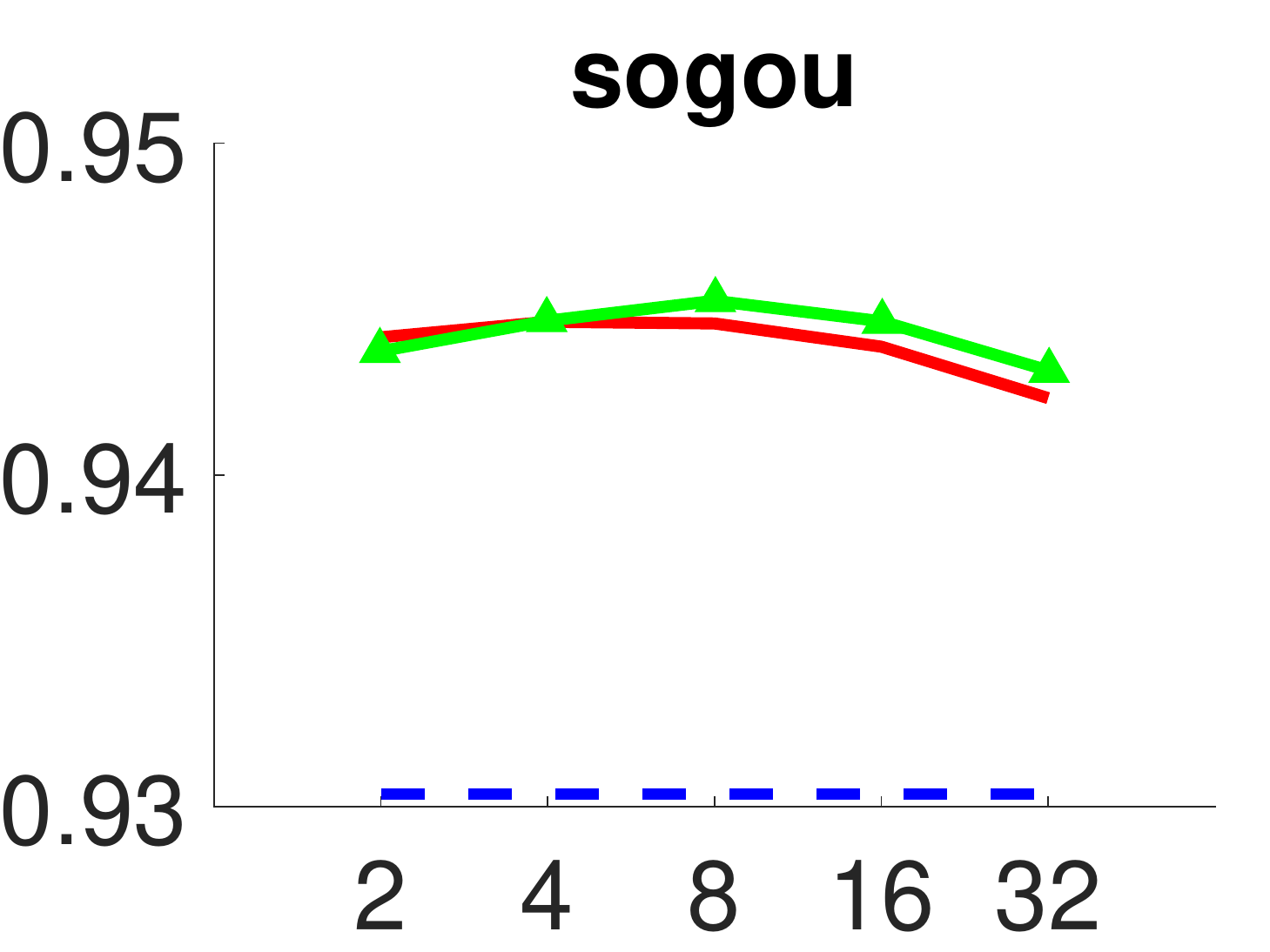}} \\
\centering
\subfloat[]{\label{fig:accu_dbpedia} \includegraphics[width 
= 0.33\linewidth]{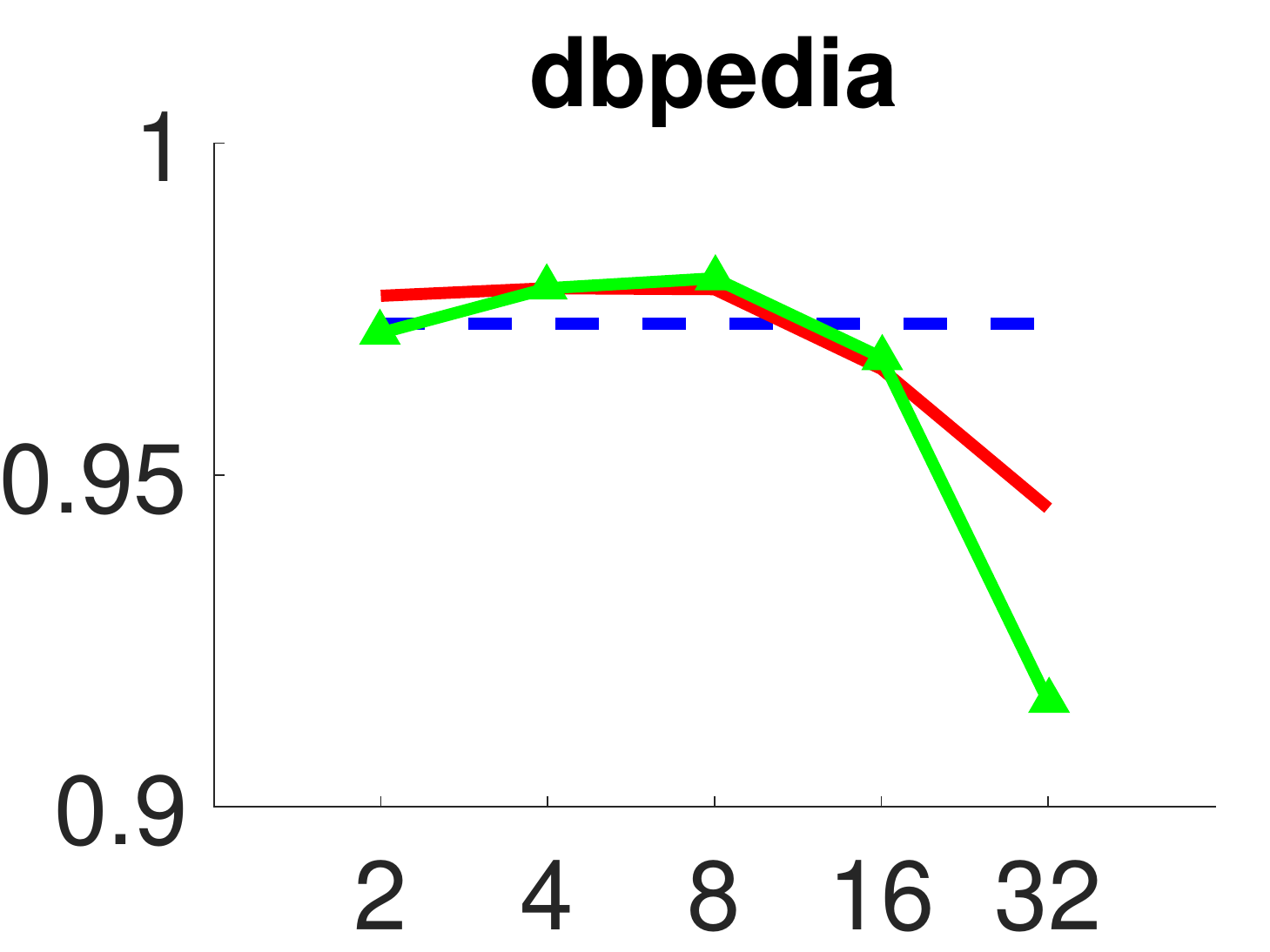}}
\centering
\subfloat[]{\label{fig:accu_yelpp} \includegraphics[width 
= 0.33\linewidth]{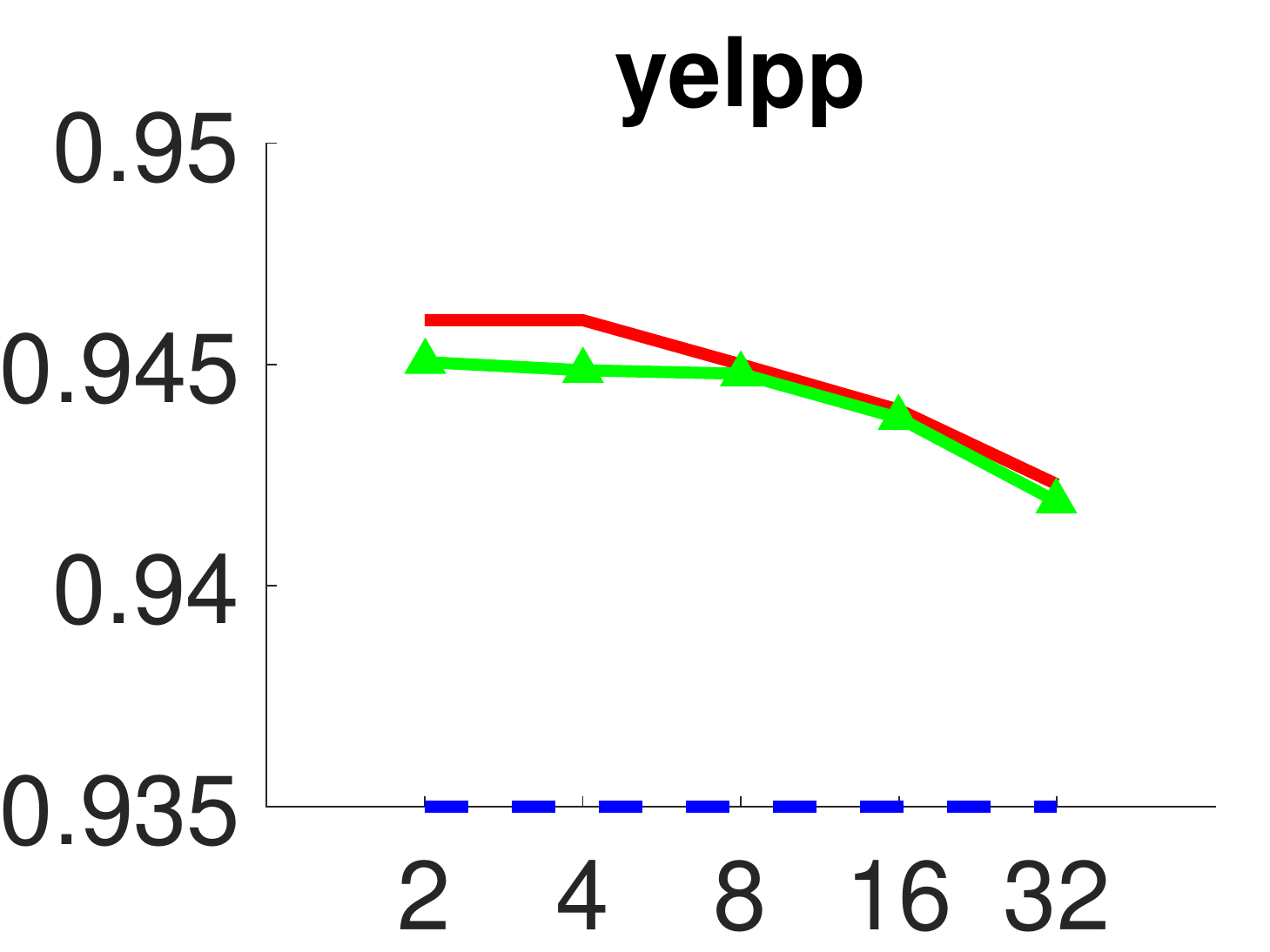}}
\centering
\subfloat[]{\label{fig:accu_yelpf} \includegraphics[width 
= 0.33\linewidth]{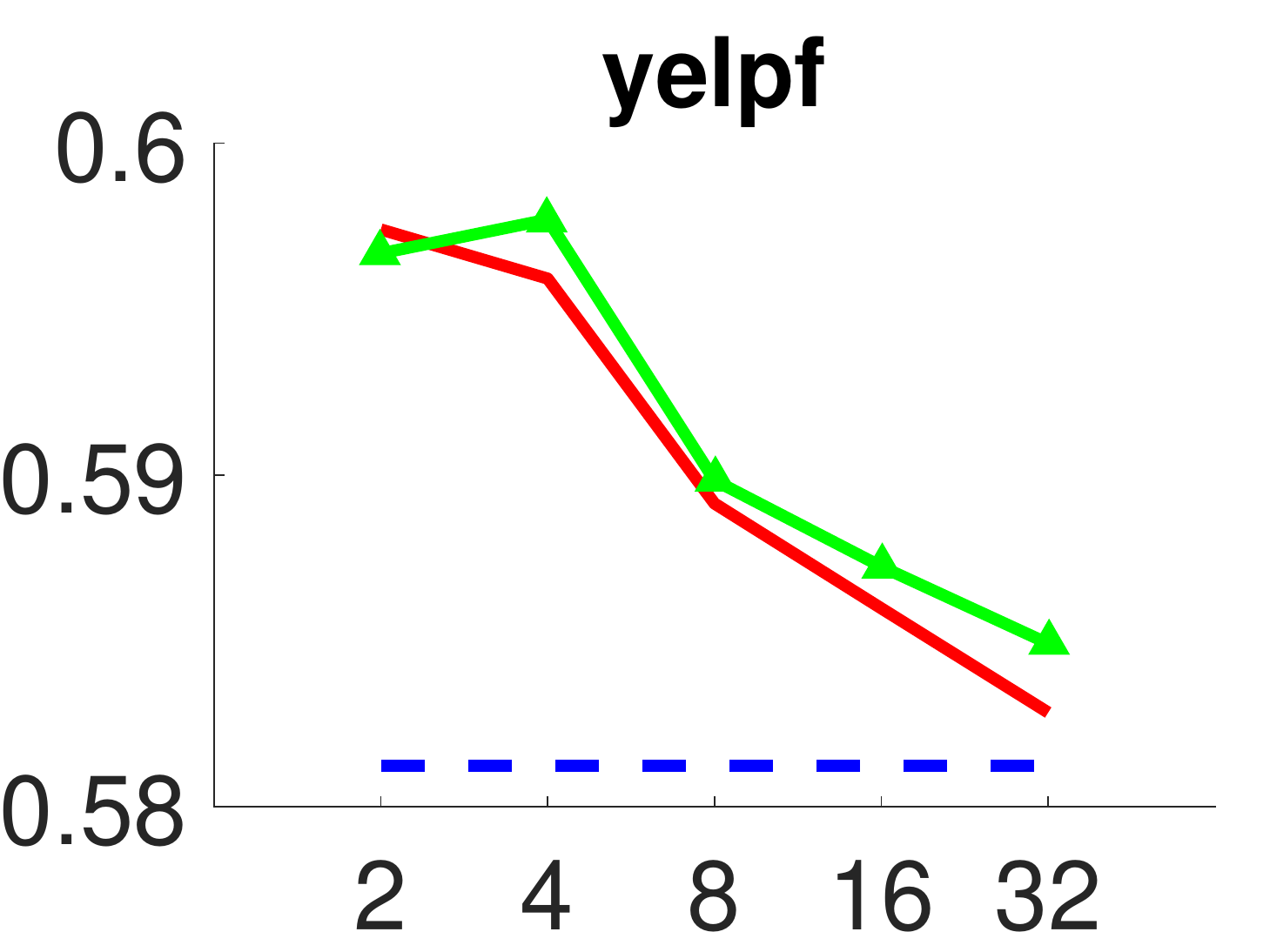}} \\
\centering
\subfloat[]{\label{fig:accu_yahoo} \includegraphics[width 
= 0.33\linewidth]{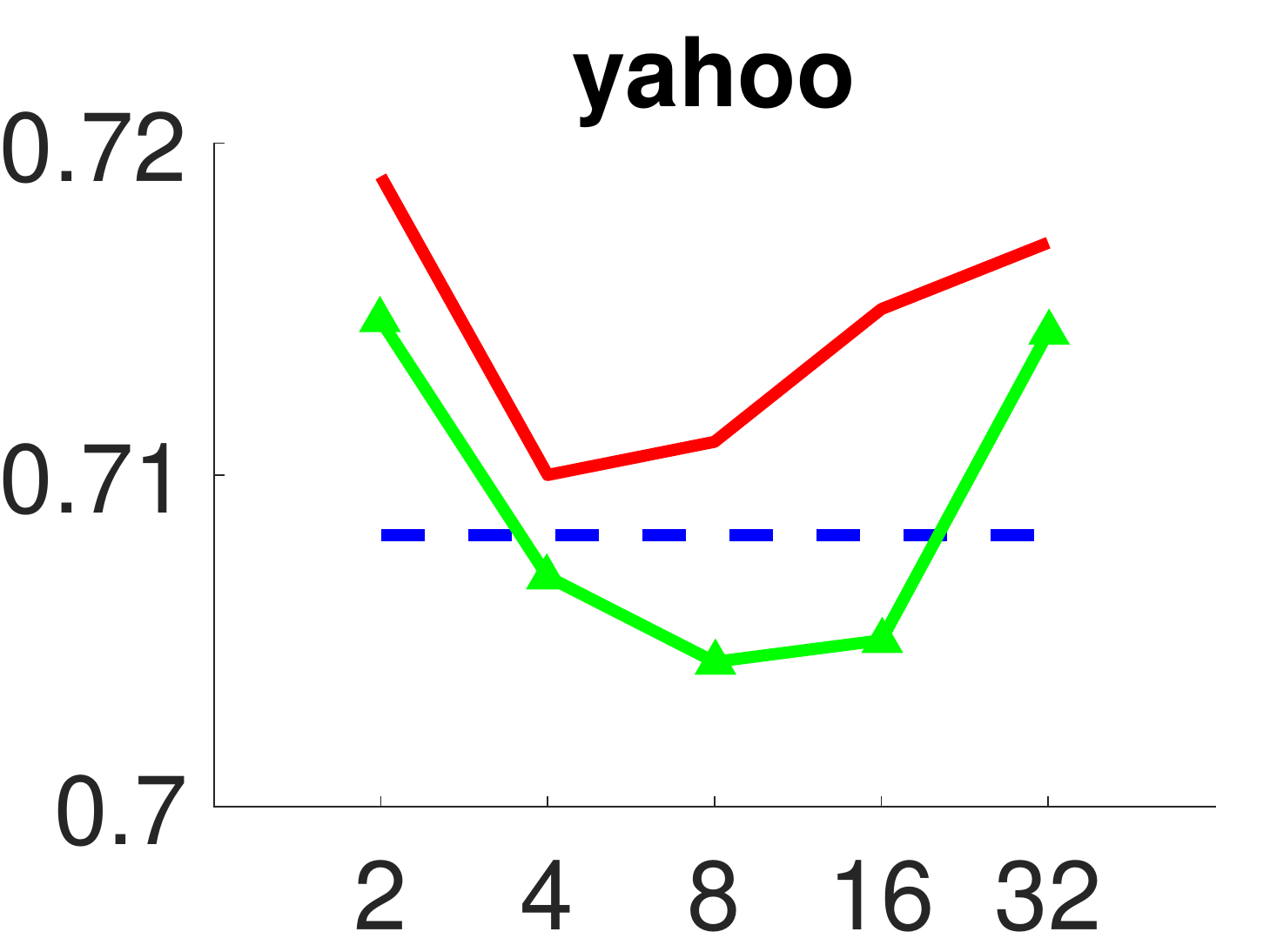}}
\centering
\subfloat[]{\label{fig:accu_amzp} \includegraphics[width 
= 0.33\linewidth]{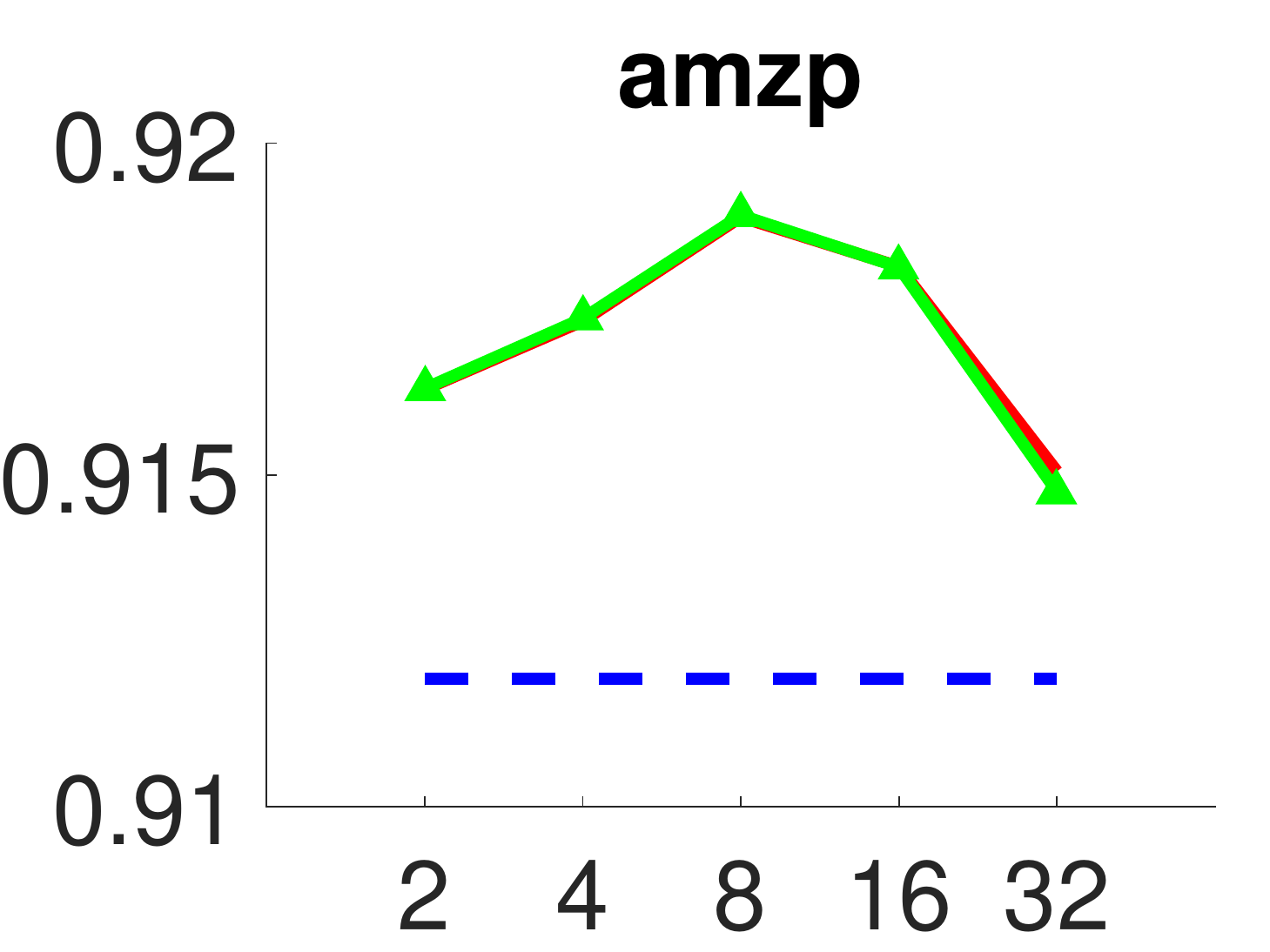}}
\centering
\subfloat[]{\label{fig:accu_amzf} \includegraphics[width 
= 0.33\linewidth]{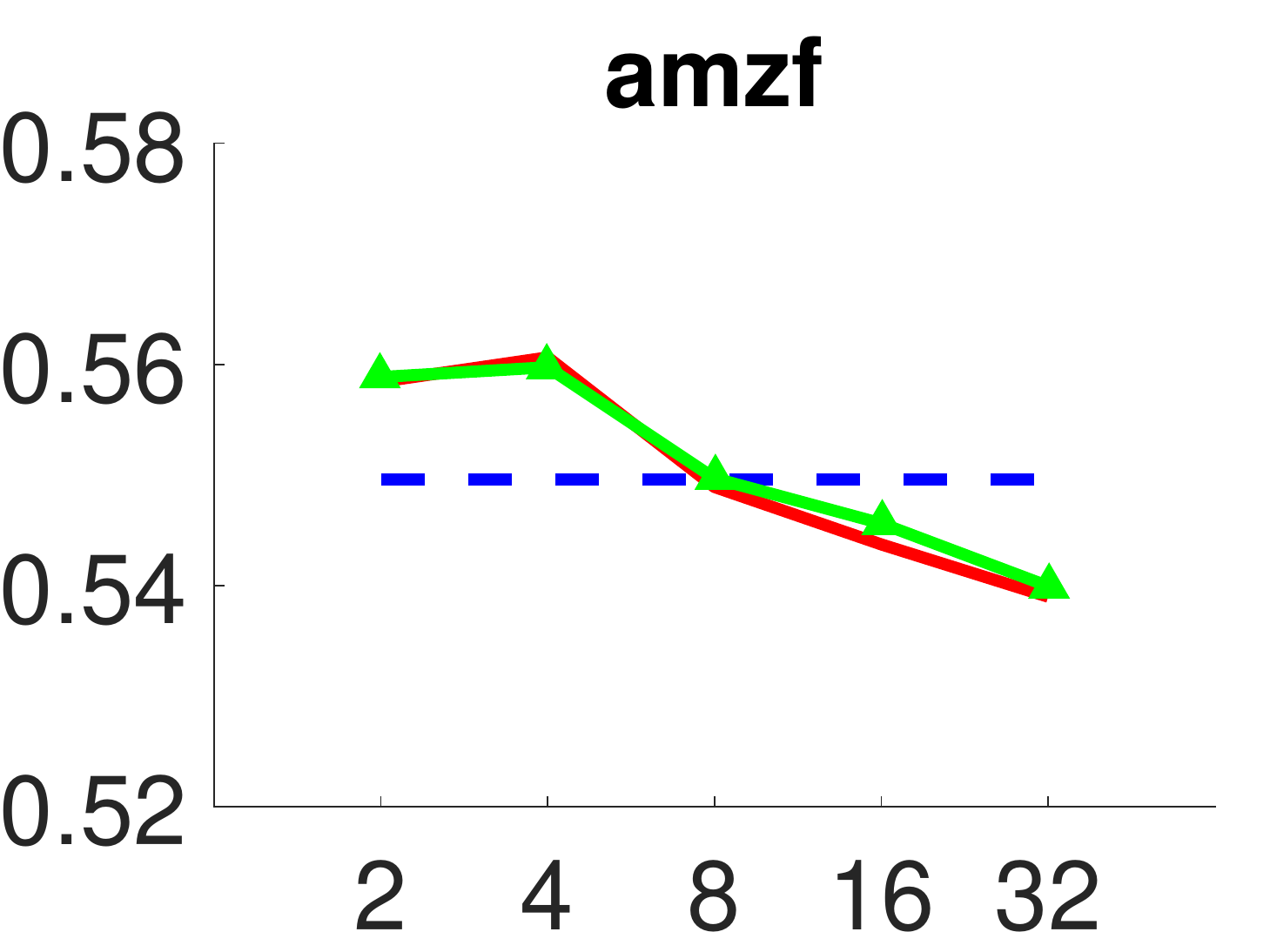}} 
\caption{Comparison of testing accuracy (\%) of fastText (dotted blue),
TMPCA+Dense (red solid), and PCA+Dense (green head dotted), where the
horizontal axis is the input length $N$. }\label{fig:exp_accu}
\end{figure}

To show the information preservation property of TMPCA, we include
fastText in the comparison. Since the difference between these three
models is the way to compute the hidden vector, we compare TMPCA, mean operation (used by fastText), and PCA.  We show the accuracy for input sequences of length 2, 4, 8, 16 ad 32 in Fig.  \ref{fig:exp_accu}. They correspond to the 1-, 2-, 3-, 4- and 5-stage TMPCA, respectively.  We show two relative mutual information values in Table \ref{tab:exp_mi_tmpca} and Table \ref{tab:exp_mi_pca}.  Table \ref{tab:exp_mi_tmpca} provides the mutual information ratio between TMPCA and mean.  Table
\ref{tab:exp_mi_pca} offers the mutual information ratio between PCA and
TMPCA. We see that TMPCA is much more capable than mean and is
comparable with PCA in preserving the mutual information.  Although
higher mutual information does not always translate into better
classification performance, there is a strong correlation between them.
This substantiates our mutual information discussion.  We should point
out that the mutual information on different inputs (in our case,
different $N$ values) is not directly comparable. Thus, a higher
relative mutual information value on longer inputs cannot be interpreted
as containing richer information and, consequently, higher accuracy. We
observe that the dense network achieves its best performance when $N=4$
or $8$. 

\begin{figure}[htbp]
\hspace*{-0.1in}
\centering
\subfloat[]{\label{fig:energy_spam} \includegraphics[width 
= 0.33\linewidth]{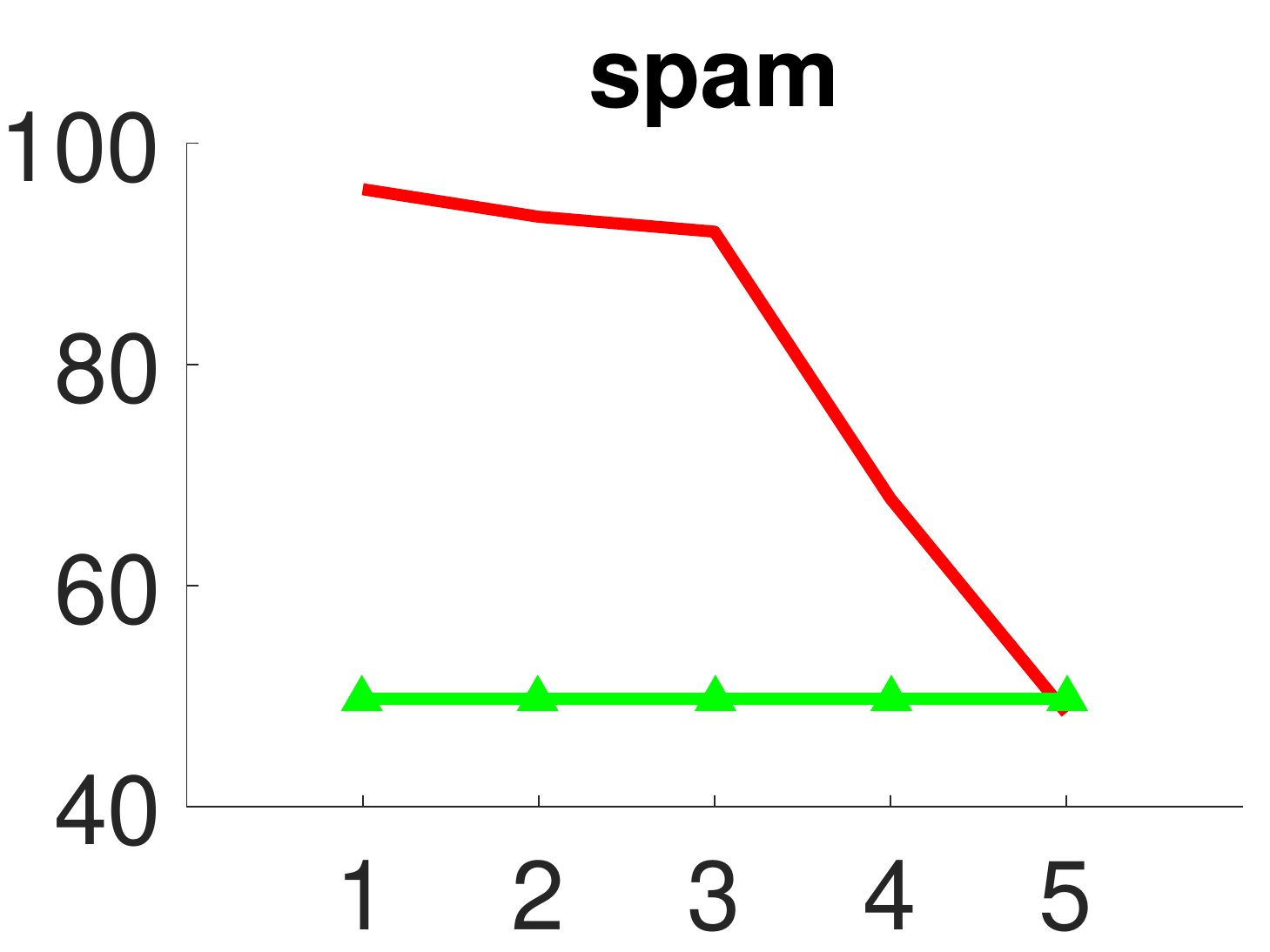}}
\centering
\subfloat[]{\label{fig:energy_sst} \includegraphics[width 
= 0.33\linewidth]{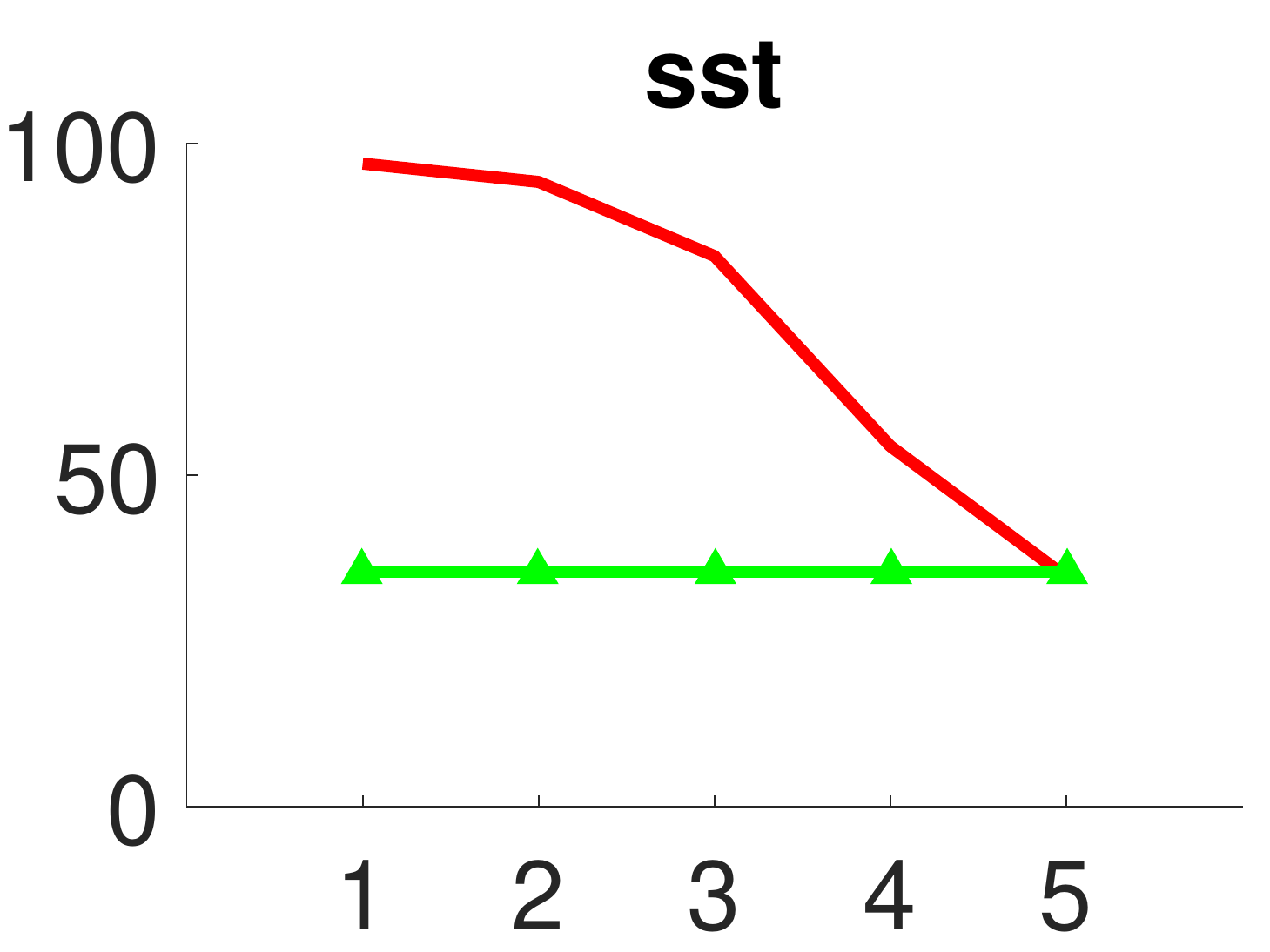}}
\centering
\subfloat[]{\label{fig:energy_semeval} \includegraphics[width 
= 0.33\linewidth]{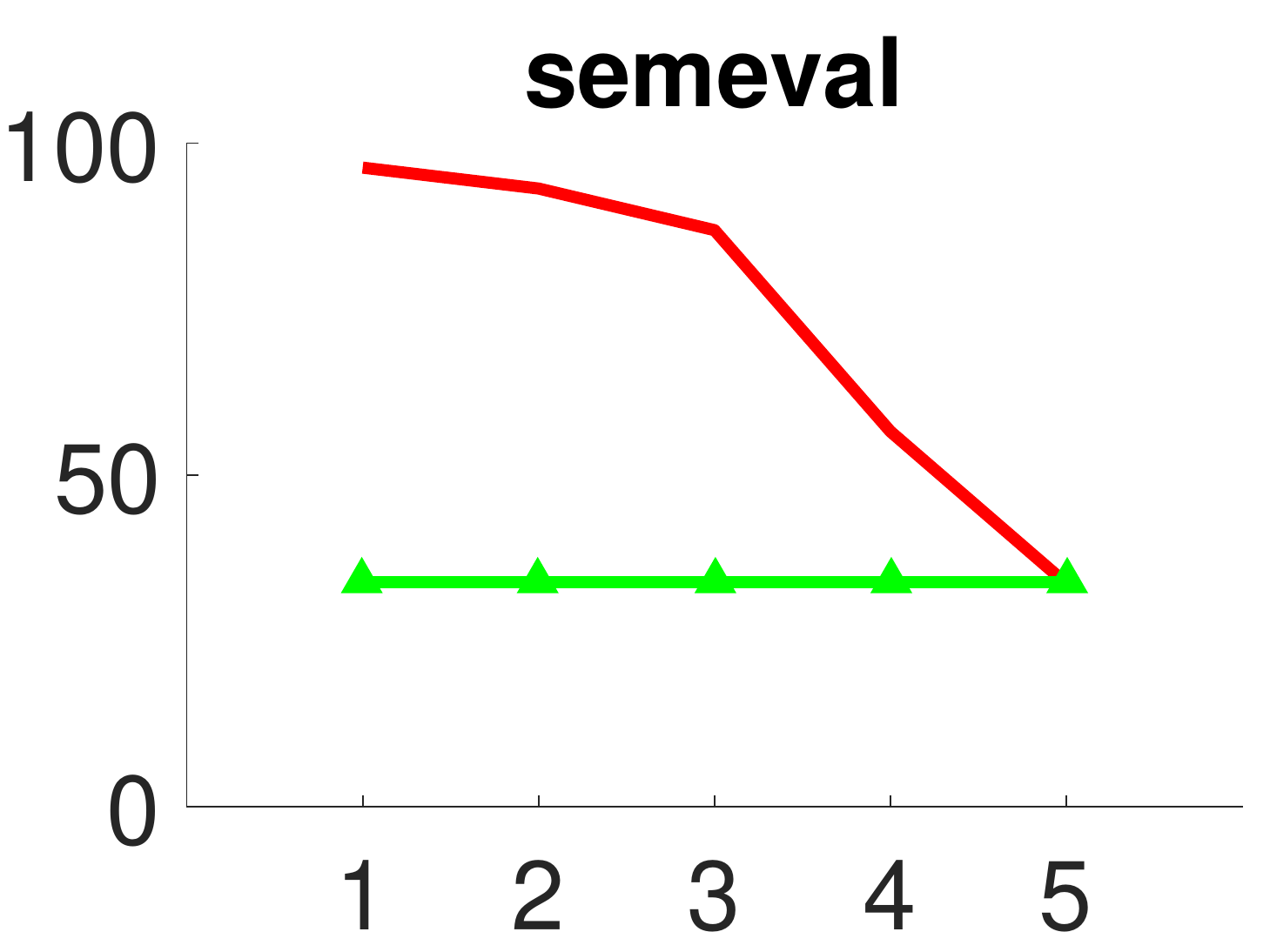}} \\
\centering
\subfloat[]{\label{fig:energy_imdb} \includegraphics[width 
= 0.33\linewidth]{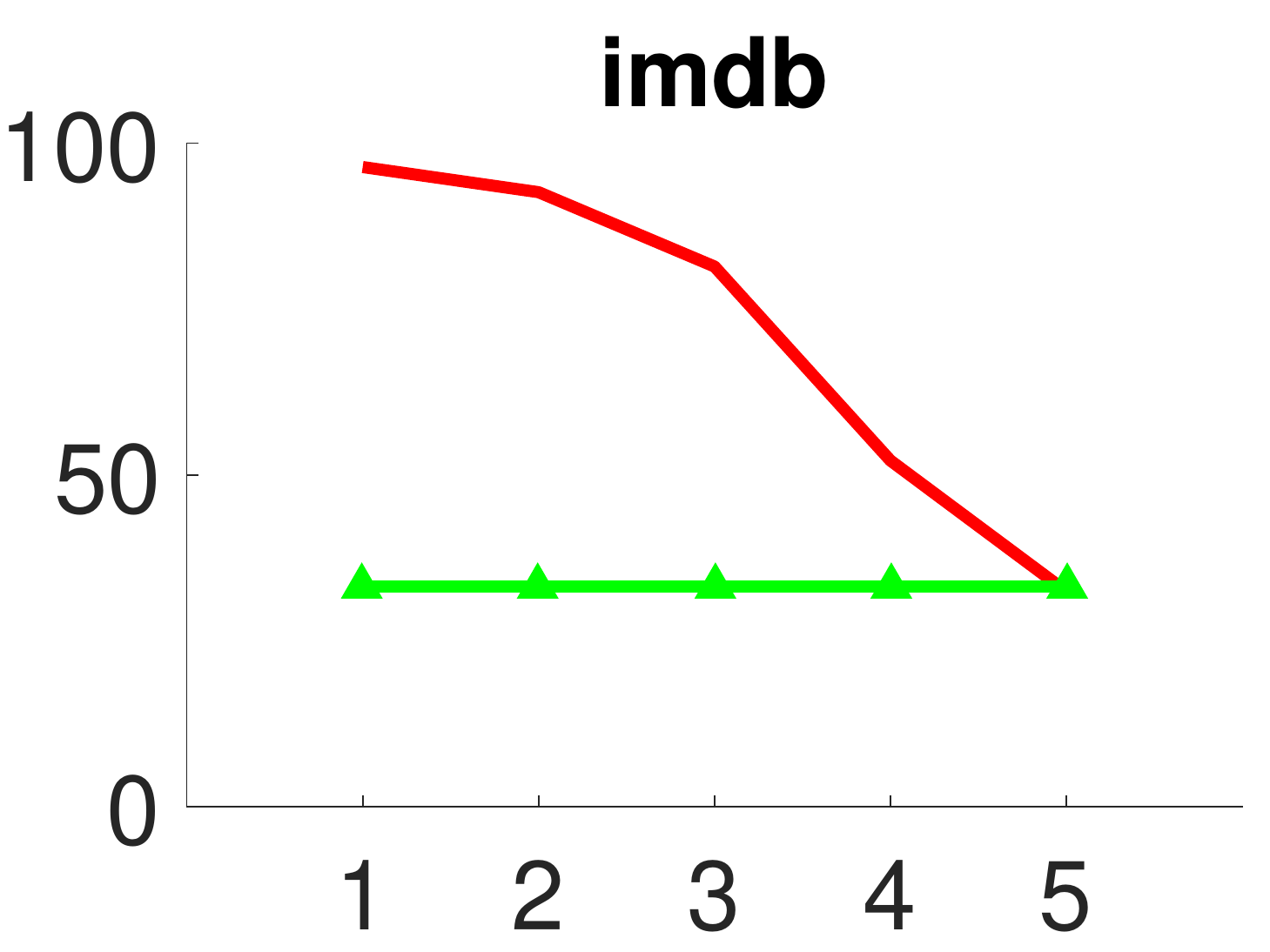}}
\centering
\subfloat[]{\label{fig:energy_agnews} \includegraphics[width 
= 0.33\linewidth]{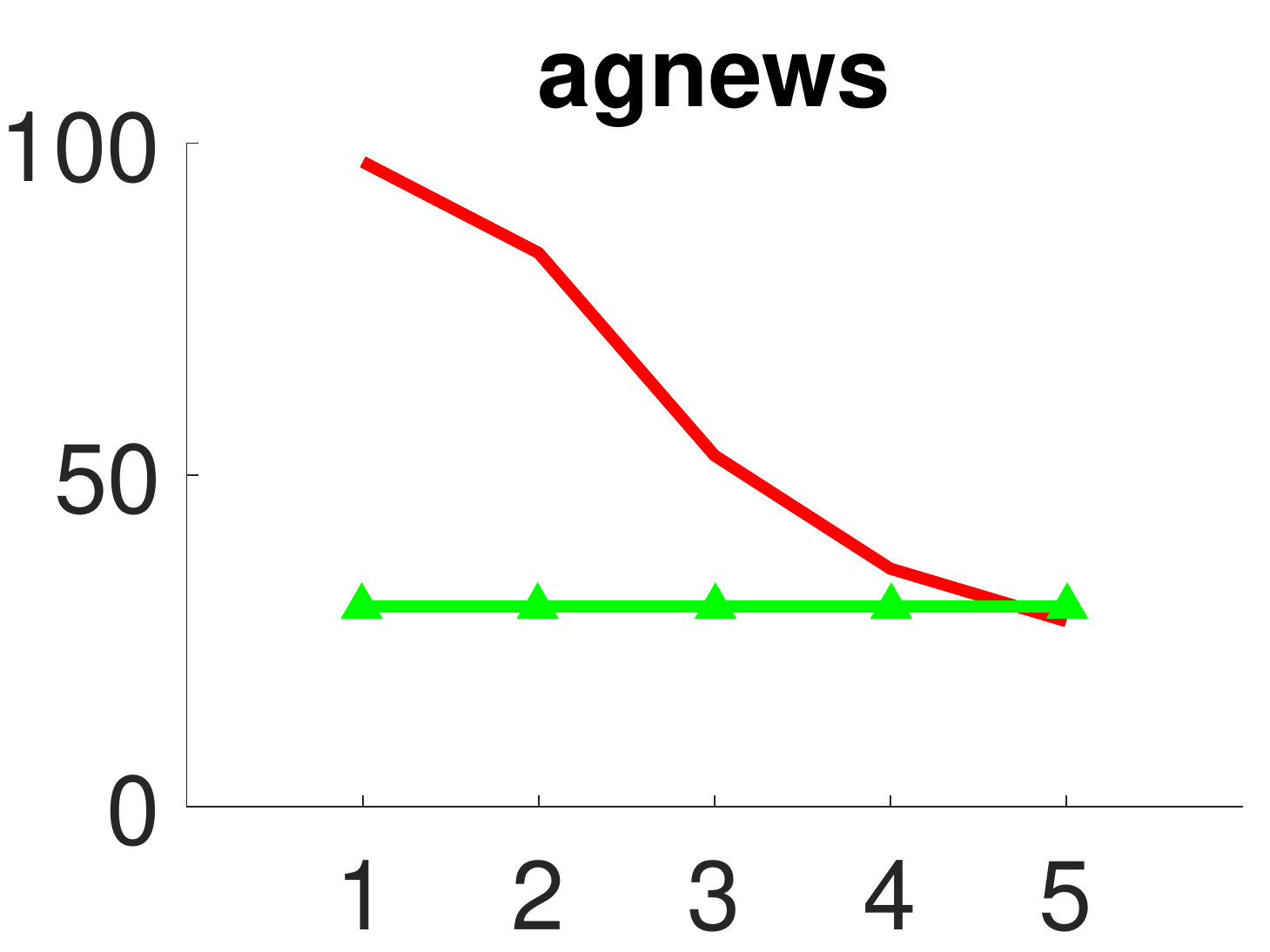}}
\centering
\subfloat[]{\label{fig:energy_sogou} \includegraphics[width 
= 0.33\linewidth]{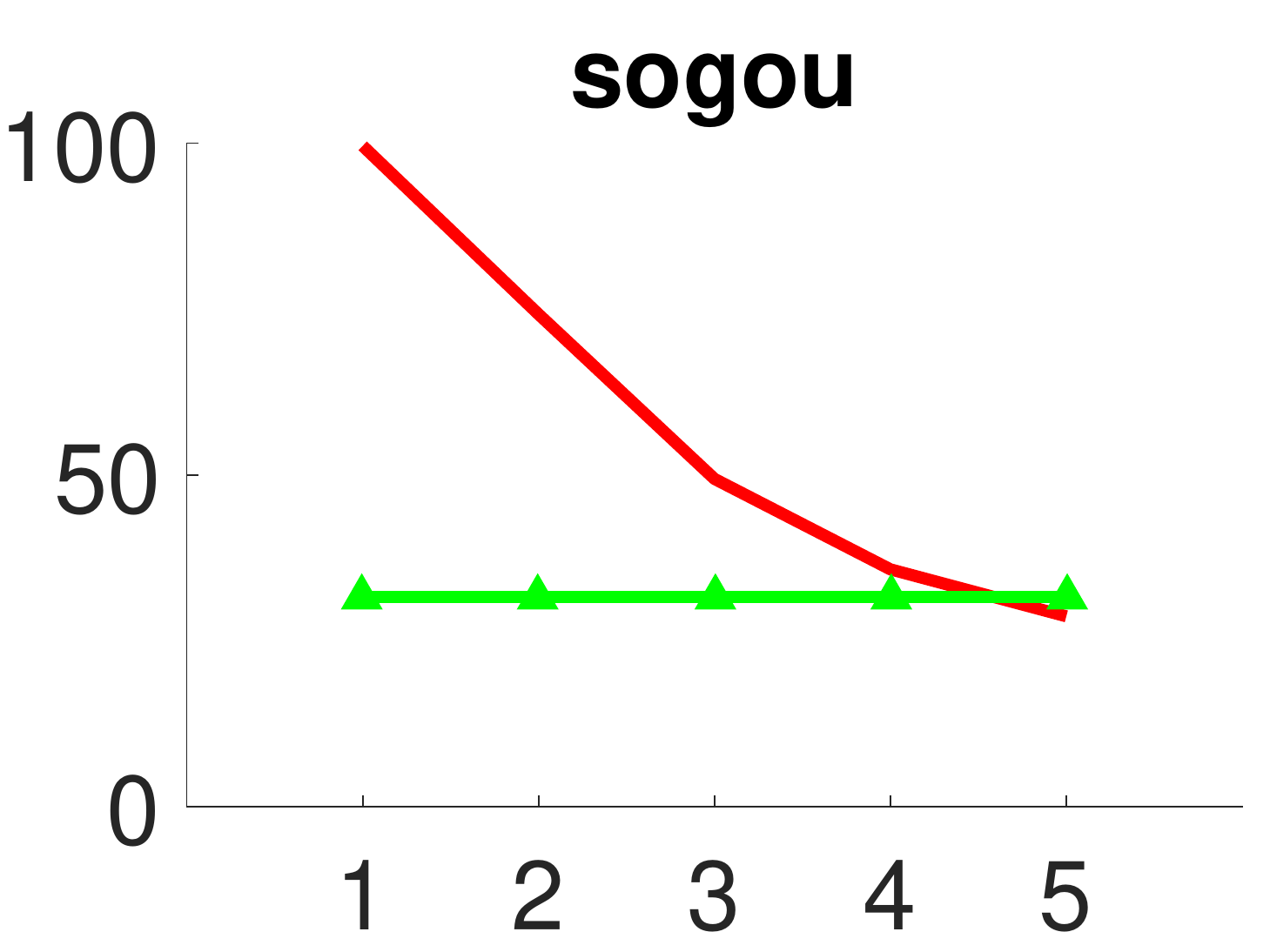}} \\
\centering
\subfloat[]{\label{fig:energy_dbpedia} \includegraphics[width 
= 0.33\linewidth]{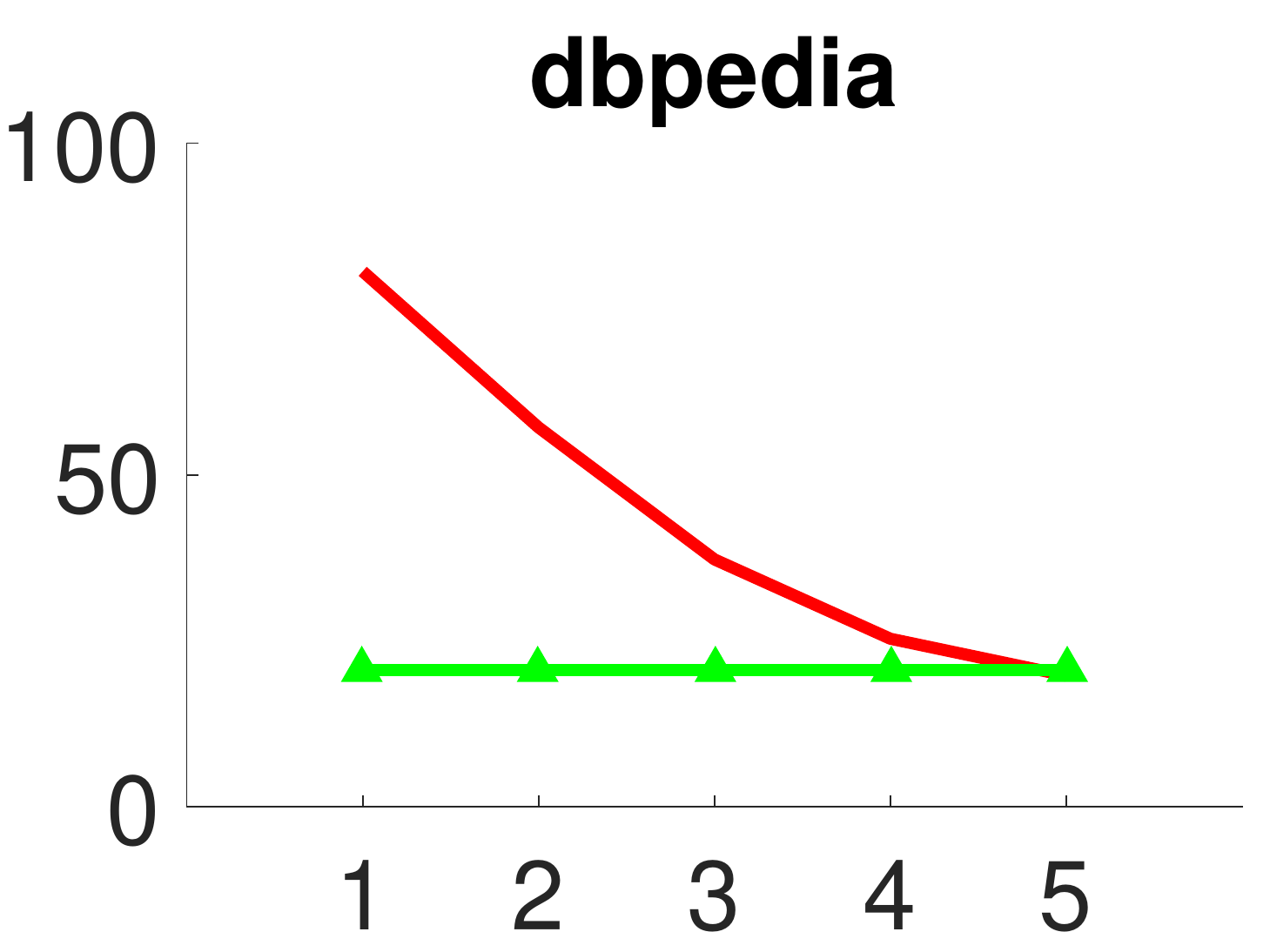}}
\centering
\subfloat[]{\label{fig:energy_yelpp} \includegraphics[width 
= 0.33\linewidth]{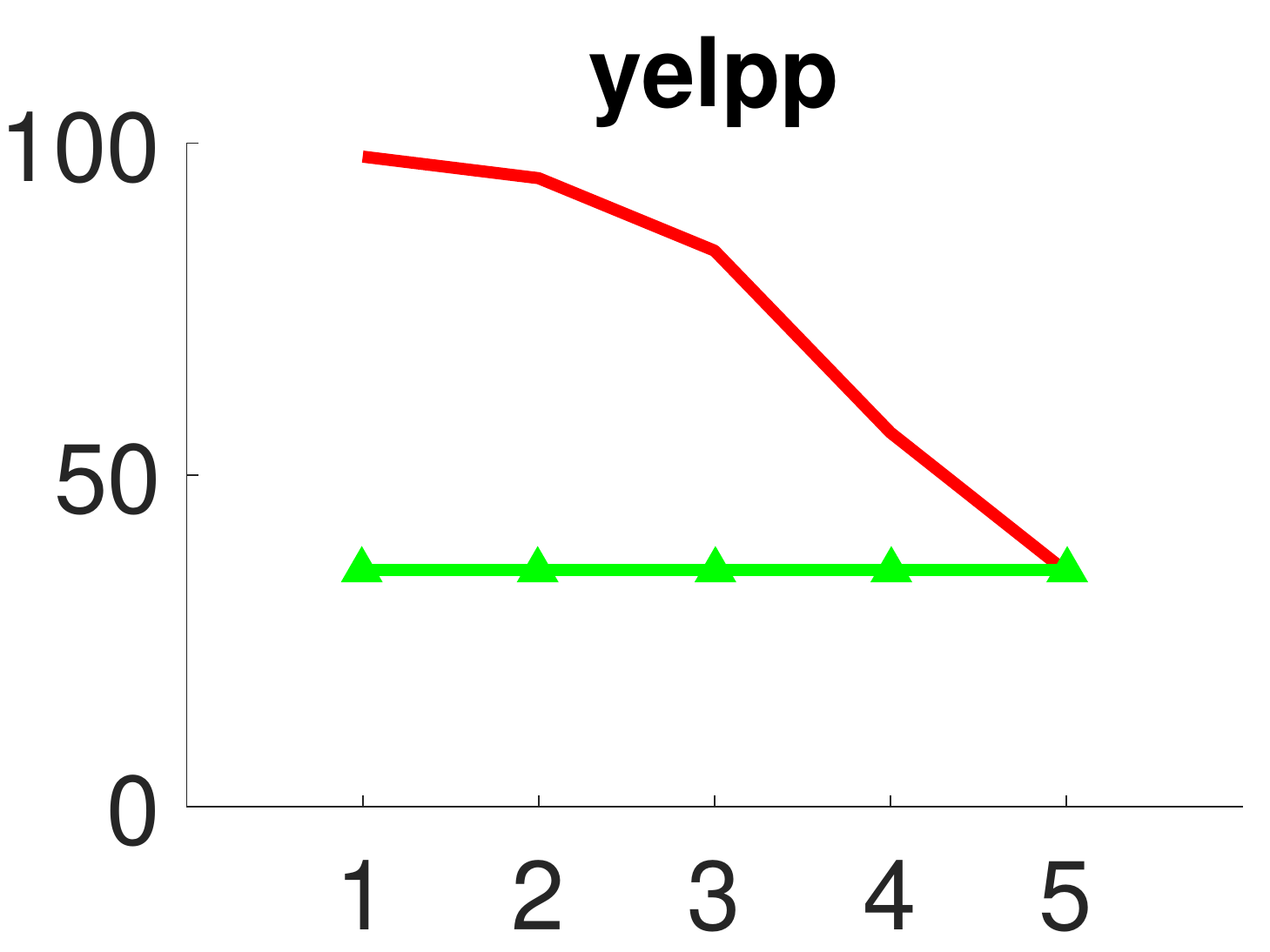}}
\centering
\subfloat[]{\label{fig:energy_yelpf} \includegraphics[width 
= 0.33\linewidth]{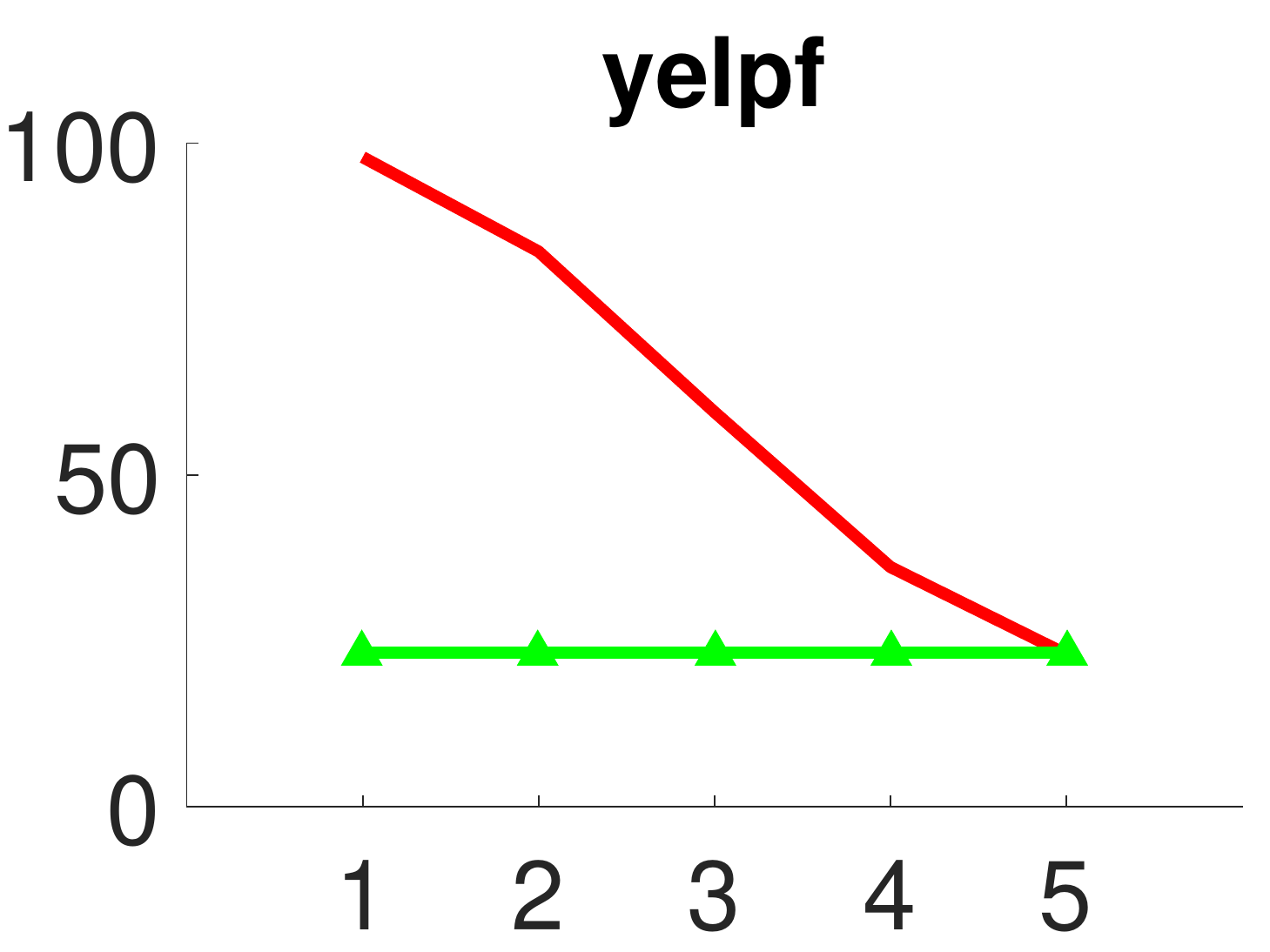}} \\
\centering
\subfloat[]{\label{fig:energy_yahoo} \includegraphics[width 
= 0.33\linewidth]{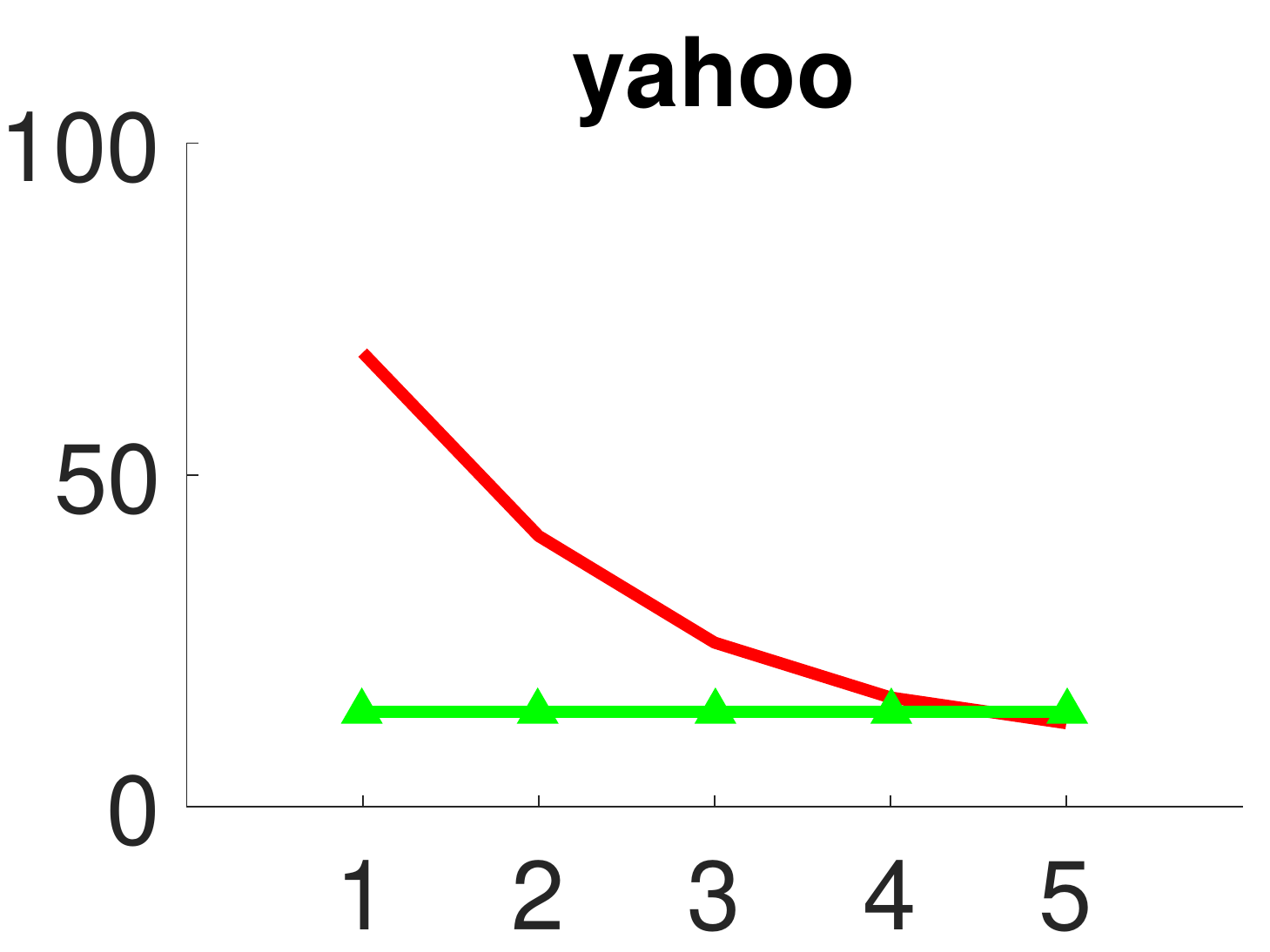}}
\centering
\subfloat[]{\label{fig:energy_amzp} \includegraphics[width 
= 0.33\linewidth]{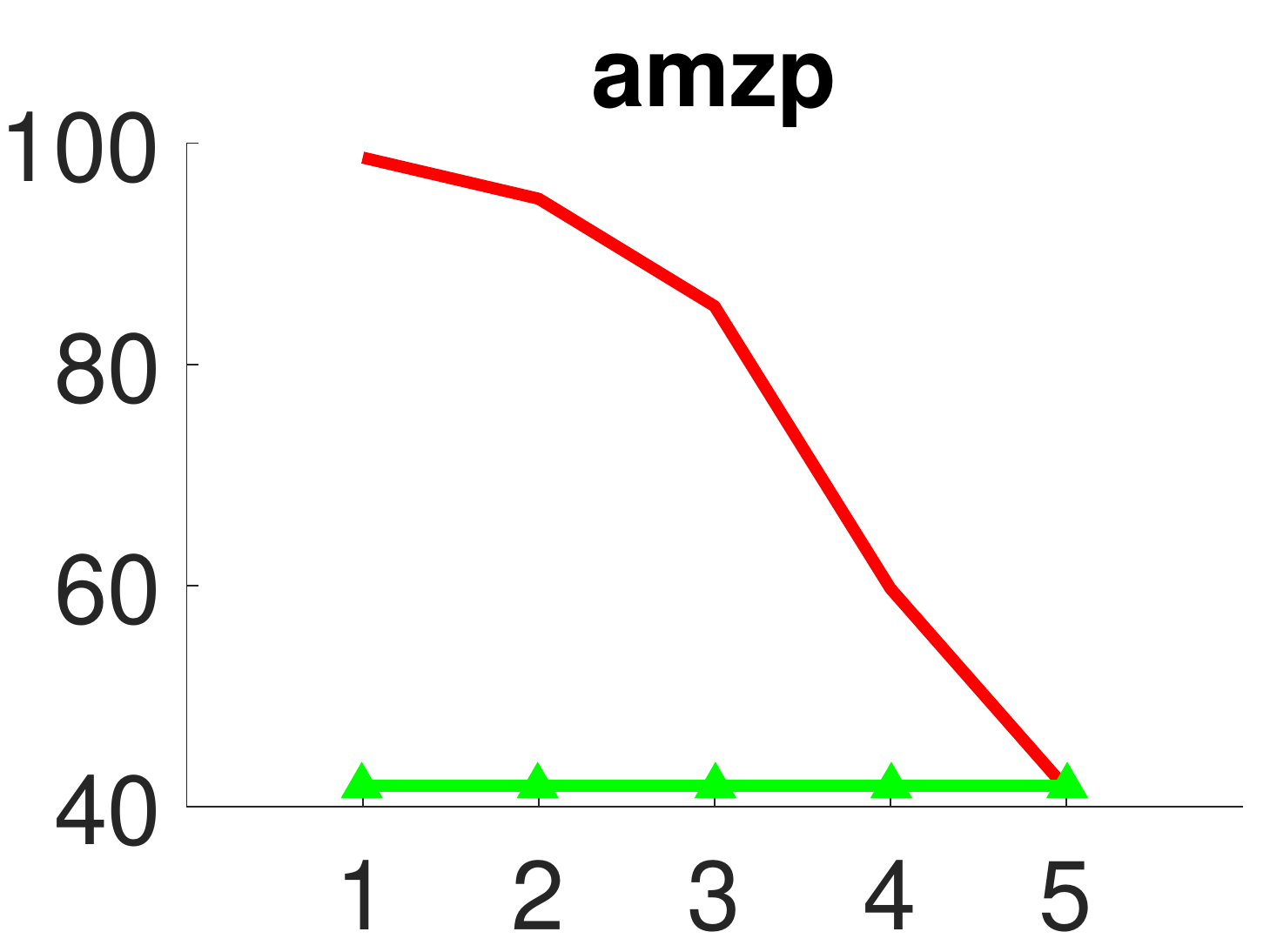}}
\centering
\subfloat[]{\label{fig:energy_amzf} \includegraphics[width 
= 0.33\linewidth]{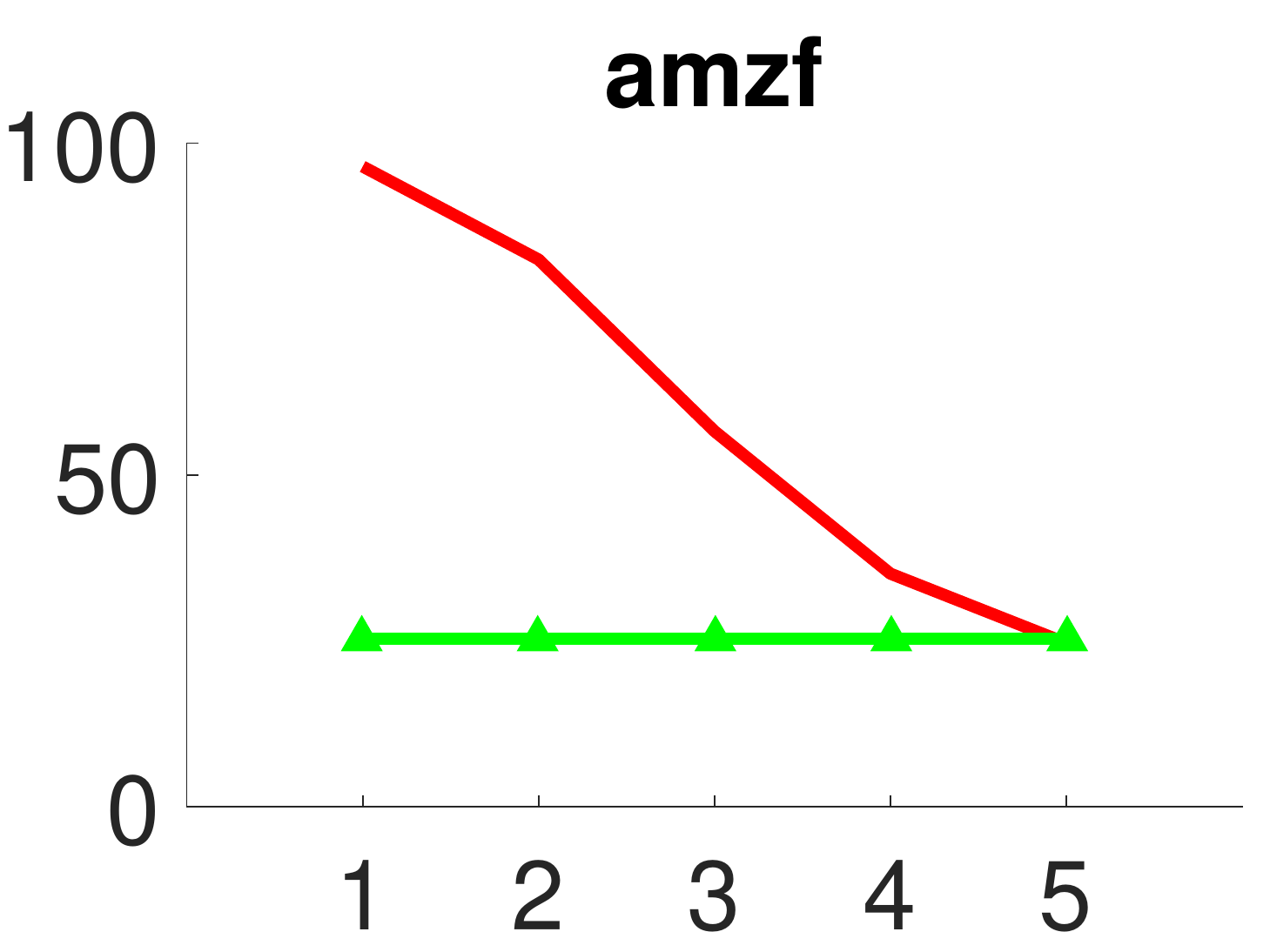}}
\caption{The energy of TMPCA (red solid) and PCA (green head dotted)
coefficients is expressed as percentages of the energy of input
sequences of length $N=32$, where the horizontal axis indicates the
TMPCA stage number while PCA has only one stage.} \label{fig:exp_energy}
\end{figure}

To understand information loss at each TMPCA, we plot their energy
percentages in Fig. \ref{fig:exp_energy} where the input has a length of
$N=32$. For TMPCA, the energy drops as the number of stage increases,
and the sharp drop usually happens after 2 or 3 stages.  This
observation is confirmed by the results in Fig.  \ref{fig:exp_accu}.
For performance benchmarking, we provide the energy percentage of PCA in
the same figure. Since the PCA has only one stage, we use a horizontal
line to represent the percentage level. Its value is equal or slightly
higher than the energy percentage at the final stage of TMPCA. This is
collaborated by the closeness of their mutual information values in
Table \ref{tab:exp_mi_pca}.  The information preserving and the low
computational complexity properties make TMPCA an excellent dimension
reduction pre-processing tool for text classification. 

%%%%%%%%%%%%%%%%%%%%%%%%%%%%%%%%%%%%%%%%%%%%%%%%%%%%%%%%%%%%%
\begin{table*}[htbp]
\centering
\caption{Comparison of training time in seconds (TMPCA/PCA).}\label{tab:exp_time}
\begin{tabular*}{\textwidth}{@{\extracolsep{\fill} }l c c c c} \hline
     & $N=4$ & $N=8$ & $N=16$ & $N=32$\\ \hline
spam & \textbf{0.007}/0.023& \textbf{0.006}/0.090& \textbf{0.007}/0.525& \textbf{0.011}/7.389\\ \hline 
sst & \textbf{0.007}/0.023& \textbf{0.006}/0.090& \textbf{0.008}/0.900& \textbf{0.009}/5.751\\ \hline 
semeval & \textbf{0.005}/0.017& \textbf{0.007}/0.111& \textbf{0.021}/2.564& \textbf{0.009}/5.751\\ \hline 
imdb&  \textbf{0.006}/0.019& \textbf{0.008}/0.114& \textbf{0.009}/0.781& \textbf{0.009}/6.562\\ \hline 
agnews&  \textbf{0.014}/0.053& \textbf{0.017}/0.325& \textbf{0.033}/4.100& \textbf{0.061}/47.538\\ \hline 
sogou&  \textbf{0.029}/0.111& \textbf{0.053}/1.093& \textbf{0.134}/17.028& \textbf{0.214}/173.687\\ \hline 
dbpedia&  \textbf{0.039}/0.145& \textbf{0.092}/1.886& \textbf{0.125}/15.505& \textbf{0.348}/279.405\\ \hline 
yelpp&  \textbf{0.037}/0.145& \textbf{0.072}/1.517& \textbf{0.163}/20.740& \textbf{0.272}/222.011\\ \hline 
yelpf &  \textbf{0.035}/0.137& \textbf{0.072}/1.517& \textbf{0.157}/19.849& \textbf{0.328}/268.698\\ \hline 
yahoo&  \textbf{0.068}/0.269& \textbf{0.129}/2.714& \textbf{0.322}/40.845& \textbf{0.787}/642.278\\ \hline 
amzp & \textbf{0.184}/0.723& \textbf{0.379}/8.009& \textbf{0.880}/112.021& \textbf{1.842}/1504.912\\ \hline 
amzf& \textbf{0.167}/0.665& \textbf{0.351}/7.469& \textbf{0.778}/99.337& \textbf{1.513}/1237.017\\ \hline 
\end{tabular*}
\end{table*}
%%%%%%%%%%%%%%%%%%%%%%%%%%%%%%%%%%%%%%%%%%%%%%%%%%%%%%%%%%%%%

%%%%%%%%%%%%%%%%%%%%%%%%%%%%%%%%%%%%%%%%%%%%%%%%%%%%%%%%%%%%%
%%%%%%%%%%%%%%%%%%%%%%%%%%%%%%%%%%%%%%%%%%%%%%%%%%%%%%%%%%%%%

%%%%%%%%%%%%%%%%%%%%%%%%%%%%%%%%%%%%%%%%%%%%%%%%%%%%%%%%%%%%%
\begin{table*}[htbp]
\centering
\caption{The relative mutual information ratio (TMPCA versus Mean).}
\label{tab:exp_mi_tmpca}
\begin{tabular*}{\textwidth}{@{\extracolsep{\fill} }l c c c c c} \hline
& $N=2$ & $N=4$ & $N=8$ & $N=16$ & $N=32$\\ \hline
spam& 1.32e+02& 7.48e+05& 2.60e+12& 5.05e+14& 9.93e+12\\ \hline 
sst& 8.48e+03& 1.22e+10& 1.28e+15& 8.89e+15& 9.17e+13\\ \hline 
semeval& 5.52e+03& 1.13e+09& 3.30e+14& 4.78e+15& 1.67e+13\\ \hline 
imdb& 1.34e+04& 3.49e+09& 1.89e+14& 8.73e+14& 1.05e+13\\ \hline 
agnews& 4.10e+05& 5.30e+10& 7.09e+11& 3.56e+12& 6.11e+12\\ \hline 
sogou& 5.53e+08& 1.37e+13& 6.74e+13& 5.40e+13& 4.21e+13\\ \hline 
dbpedia& 20.2& 111& 227& 814& 306\\ \hline 
yelpp& 8.42e+04& 2.79e+11& 3.85e+15& 5.65e+16& 1.46e+16\\ \hline 
yelpf& 2.29e+07& 1.90e+11& 5.92e+12& 5.42e+12& 1.58e+12\\ \hline 
yahoo& 6.7& 9.1& 9.9& 5.8& 1.5\\ \hline 
amzp& 7.34e+05& 4.48e+11& 1.24e+16& 1.15e+18& 2.75e+18\\ \hline 
amzf& 3.09e+06& 1.47e+10& 3.38e+11& 1.48e+12& 2.37e+12\\ \hline 
\end{tabular*}
\end{table*}
%%%%%%%%%%%%%%%%%%%%%%%%%%%%%%%%%%%%%%%%%%%%%%%%%%%%%%%%%%%%%

%%%%%%%%%%%%%%%%%%%%%%%%%%%%%%%%%%%%%%%%%%%%%%%%%%%%%%%%%%%%%
\begin{table*}[htbp]
\centering
\caption{The relative mutual information ratio (PCA versus TMPCA).}
\label{tab:exp_mi_pca}
\begin{tabular*}{\textwidth}{@{\extracolsep{\fill} }l c c c c}
\hline
& $N=4$ & $N=8$ & $N=16$ & $N=32$\\ \hline
spam&1.04& 1.00& 1.00& 1.49\\ \hline 
sst&  1.00& 1.00& 1.00& 1.36\\ \hline 
semeval&  0.99& 1.00& 1.00& 1.09\\ \hline 
imdb& 1.02& 1.00& 1.00& 1.29\\ \hline 
agnews&  1.00& 1.01& 1.40& 2.92\\ \hline 
sogou& 1.00& 1.20& 1.66& 5.17\\ \hline 
dbpedia&  1.16& 1.63& 1.65& 1.75\\ \hline 
yelpp&1.00& 1.00& 1.00& 1.13\\ \hline 
yelpf& 1.00& 1.01& 1.01& 1.10\\ \hline 
yahoo&  1.01& 1.30& 1.94& 8.78\\ \hline 
amzp&  1.00& 1.00& 1.00& 1.10\\ \hline 
amzf&  1.00& 1.00& 1.03& 1.41\\ \hline 
\end{tabular*}
\end{table*}
%%%%%%%%%%%%%%%%%%%%%%%%%%%%%%%%%%%%%%%%%%%%%%%%%%%%%%%%%%%%%

%%%%%%%%%%%%%%%%%%%%%%%%%%%%%%%%%%%%%%%%%%%%%%%%%%%%%%%%%%%%%
%%%%%%%%%%%%%%%%%%%%%%%%%%%%%%%%%%%%%%%%%%%%%%%%%%%%%%%%%%%%%

\section{Conclusion}\label{sec:con} 

An efficient language data dimension reduction technique, called the
TMPCA method, was proposed for TC problems in this work. TMPCA is a
multi-stage PCA in special form, and it can be described by a transform
matrix with orthonormal rows. It can retain the input information by
maximizing the mutual information between its input and output, which is
beneficial to TC problems. It was shown by experimental results that a
dense network trained on the TMPCA preprocessed data outperforms
state-of-the-art fastText, char-CNN and LSTM in quite a few TC datasets.
Furthermore, the number of parameters used by TMPCA is an order of
magnitude smaller than other NN-based models.  Typically, TMPCA takes
less than one second training time on a large-scale dataset that has
millions of samples. To conclude, TMPCA is a powerful dimension
reduction pre-processing tool for text classification for its low
computational complexity, low storage requirement for model parameters
and high information preserving capability. 

\section{Declarations of interest}\label{sec:declarations}
Declarations of interest: none

\section{Acknowledgements}\label{sec:acknowledgments}
This research did not receive any specific grant from funding agencies in 
the public, commercial, or not-for-profit sectors.

\section*{Appendix A: Detailed Derivation of TMPCA System Function}\label{app_TMPCA_func}

We use the same notations in Sec. \ref{sec:TMPCA}. For stage $s>1$, we have:
\begin{align}\label{eq:TMPCA_func_1}
z^s_j = U^s \begin
		{bmatrix} z^{s-1}_{2j-1}\\ 
		z^{s-1}_{2j}\end{bmatrix} 
      = U^s_1 z^{s-1}_{2j-1} + U^s_2 z^{s-1}_{2j}, 
\end{align} 
where $j = 1, \cdots, \frac{N}{2^s}$. When $s=1$, we have 
\begin{equation}\label{eq:TMPCA_func_2}
z^1_j = U^1_1w_{2j-1}+U^1_2 w_{2j}
\end{equation}
From Eqs. (\ref{eq:TMPCA_func_1}) and (\ref{eq:TMPCA_func_2}), we get
\begin{align}\label{eq:TMPCA_func_v1}
Y &= z^L_1 = \sum^N_{j=1} \Big( \prod^L_{s=1} U^s_{f_{j,s}}\Big) w_j, \\
f_{j,s} &= b_L(j-1)_s + 1,
\end{align}
where $b_L (x)_s$ is the $s$th digit of the binarization of $x$ of
length $L$. Eq. (\ref{eq:TMPCA_func_v1}) can be further simplified to
Eq. (\ref{eq:TMPCA_func}).  For example, if $N=8$, we obtain
\begin{align}
Y = &U^3_1U^2_1U^1_1w_1 + U^3_1U^2_1U^1_2w_2 + U^3_1U^2_2U^1_1w_3 + U^3_1U^2_2U^1_2w_4 + \nonumber\\
    &U^3_2U^2_1U^1_1w_5 + U^3_2U^2_1U^1_2w_6 + U^3_2U^2_2U^1_1w_7 + U^3_2U^2_2U^1_2w_8.
\end{align}
The superscripts of $U^s_j$ are arranged in the stage order of $L,
L-1,..., 1$. The subscripts are shown in Table 
\ref{tab:TMPCA_func_subscript}. This is the reason that binarization is required to express the subscripts in Eqs.  (\ref{eq:TMPCA_func}) and 
(\ref{eq:TMPCA_func_v1}). 

%%%%%%%%%%%%%%%%%%%%%%%%%%%%%%%%%%%%%%%%%%%%%%%%%%%%%%%%%%%%%
\begin{table}[htbp]
\centering
\caption{Subscripts of $U^s_j$}\label{tab:TMPCA_func_subscript}
\begin{tabular}{ c  c  c  c  c  c  c  c } \hline
$w_1$& $w_2$ & $w_3$ & $w_4$ & $w_5$ & $w_6$ & $w_7$ & $w_8$\\ \hline
1,1,1& 1,1,2& 1,2,1& 1,2,2& 2,1,1 & 2,1,2 & 2,2,1 & 2,2,2\\ \hline
\end{tabular}
\end{table}
%%%%%%%%%%%%%%%%%%%%%%%%%%%%%%%%%%%%%%%%%%%%%%%%%%%%%%%%%%%%%

\bibliography{mybibfile}
\bibliographystyle{apacite}

\end{document}